\documentclass[sn-basic]{sn-jnl}

\usepackage{graphicx}%
\usepackage{multirow}%
\usepackage{amsmath,amssymb,amsfonts}%
\usepackage{amsthm}%
\usepackage{mathrsfs}%
\usepackage[title]{appendix}%
\usepackage{xcolor}%
\usepackage{textcomp}%
\usepackage{manyfoot}%
\usepackage{booktabs}%
\usepackage{algorithm}%
\usepackage{algorithmicx}%
\usepackage{algpseudocode}%
\usepackage{listings}%

\usepackage{makecell}
\usepackage{csquotes}
\usepackage{todonotes}
\usepackage{verbatim}
\usepackage{threeparttable}
\usepackage[justification=centering]{caption}

\newenvironment{ruletable}
{%
  \begin{center}
  \small
  \begin{tabular}{@{}c@{}}
}
{%
  \end{tabular}
  \end{center}
}

\newcommand{\tableleft}[1]{#1}
\newcommand{\tablecenter}[1]{#1}
\newcommand{\tableright}[1]{#1}

\newcommand{\imgwidth}{0.9\linewidth}

\begin{document}

\title{Toward Robust Legal Text Formalization into Defeasible Deontic Logic using LLMs}


\author[1]{\fnm{Elias} \sur{Horner}}\email{elias.horner@tuwien.ac.at}
\author[2]{\fnm{Cristinel} \sur{Mateis}}\email{cristinel.mateis@ait.ac.at}
\author[3,4]{\fnm{Guido} \sur{Governatori}}\email{g.governatori@cqu.edu.au}
\author[1]{\fnm{Agata} \sur{Ciabattoni}}\email{agata@logic.at}
    
\affil[1]{\orgname{TU Wien}, \orgaddress{\city{Vienna}, \country{Austria}}}
\affil[2]{\orgname{AIT Austrian Institute of Technology}, \orgaddress{\city{Vienna}, \country{Austria}}}
\affil[3]{\orgdiv{School of Engineering and Technology}, \orgname{Central Queensland University}, \orgaddress{\city{Rockhampton}, \country{Australia}}}
\affil[4]{\orgdiv{Artificial Intelligence and Cyber Futures Institute}, \orgname{Charles Sturt University}, \orgaddress{\city{Bathurst}, \country{Australia}}}

\abstract{
We present a comprehensive approach to the automated formalization of legal texts using large language models (LLMs), targeting their transformation into Defeasible Deontic Logic (DDL). Our method employs a structured pipeline that segments complex normative language into atomic snippets, extracts deontic rules, and evaluates them for syntactic and semantic coherence. We introduce a refined success metric that more precisely captures the completeness of formalizations, and a novel two-stage pipeline with a dedicated refinement step to improve logical consistency and coverage. The evaluation procedure has been strengthened with stricter error assessment, and we provide comparative results across multiple LLM configurations, including newly released models and various prompting and fine-tuning strategies. Experiments on legal norms from the Australian Telecommunications Consumer Protections Code demonstrate that, when guided effectively, LLMs can produce formalizations that align closely with expert-crafted representations, underscoring their potential for scalable legal informatics.
}

\keywords{
legal informatics, large language models, defeasible deontic logic, semantic formalization, prompt engineering, legal NLP
}

\maketitle

\section{Introduction}
\label{sec:introduction}
Automated legal reasoning has been a major endeavour of research in AI and Law. Numerous efforts have followed the seminal paper by \cite{BNA} on formally representing the British Nationality Act in Prolog. \cite{DBLP:journals/ail/Bench-CaponAAABBBBCFGGLLLMPSSTTVWW12} and \cite{ail:30years} provide an overview of key approaches in this field.
More recently, the idea of Rules as Code emerged as a general approach to speed up the development of law-based information systems to provide services to the citizens and businesses. The white paper by OECD by \cite{MohunRobertsOECD2020} emphasises the advantages of adopting automated legal reasoning and encoding legal provisions in a machine-readable format. However, the main challenge to this vision lies in the knowledge representation bottleneck. Anecdotal evidence from numerous large-scale encoding projects suggests that an experienced coder can only encode about 4 to 5 pages per day. A recent empirical study by \cite{Cristani:encoding} on legal coding confirms this rate. Furthermore, manually encoding large regulatory frameworks raises serious concerns about burnout. While a parallel encoding by a team of human coders seems a solution, an experiment by \cite{ail2022:Alice} points out that, very likely, the parts encoded by different members of the team are incompatible with each other and reconciling them requires some considerable overhead. 

The knowledge acquisition problem is exacerbated by the growing volume of legal information and the increasing complexity of regulatory environments. This has heightened the need for automated tools capable of interpreting and formalizing normative documents. Consequently, there is a pressing need for tools that can assist with encoding legal instruments.

The idea of employing NLP techniques, particularly categorical grammar-based approaches, for encoding norms was advanced by \cite{WynerPeters}. This idea was further explored in \cite{DBLP:conf/ruleml/WynerG13}, where it was tested on small-scale examples. Subsequently, \cite{DBLP:conf/aicol/DragoniVRG17} introduced a manually supervised pipeline for extracting a formal representation. This approach relied on deterministic parsing rules and yielded reasonable outcomes. However, successful extraction required numerous iterations and was susceptible to the specific format of the input. Furthermore, it did not result in a substantial reduction in the time required to create a comprehensive and fully functional encoding.

On the other hand, \cite{jurisin19:nlp} explored the use of ML-based NLP techniques for the normative encoding process. Their key findings were that these approaches required extensive training data, which was and still is not readily available. Furthermore, their performance was not comparable to the approach proposed by \cite{DBLP:conf/aicol/DragoniVRG17}.

While recent efforts have incorporated neural methods for legal information retrieval and summarisation, few have addressed the formalisation task with the granularity and formal logic representation we propose in this paper. For example, the recent work~\cite{BilliPS24} focusses on a single article from the Council Framework Decision 2002/584/JHA (European Arrest Warrant). The approach by \cite{icail25:sartor} proposes a pipeline based on LLM and Logical English (a structured controlled language) to translate the UK Highway Code into Prolog. In their experiment, they considered four rules from the code, and the LLM correctly translated three of them.

Large Language Models (LLMs) have emerged as powerful tools for understanding and generating natural language. However, their application to legal texts remains underexplored, particularly in tasks requiring semantic precision, such as converting legal norms into machine-interpretable representations.

This paper explores the feasibility and effectiveness of using LLMs to translate legal language into a formal representation. 
We compare this approach with the one proposed in \cite{DBLP:conf/aicol/DragoniVRG17}, where legal provisions are encoded in Defeasible Deontic Logic (DDL), a computational logic framework designed to reason about obligations, permissions, and prohibitions.

The target corpus is the Australian Telecommunications Consumer Protections Code (TCP Code), containing complex and hierarchical rules. This dataset was also used in \cite{DBLP:conf/aicol/DragoniVRG17}, and we compare our results with those reported there.

Our central hypothesis is that, with suitable prompting and architectural configurations, LLMs can assist in extracting semantically valid and logically coherent deontic rules from unstructured legal text.
The novel contribution of this work lies in the integration of prompt engineering techniques, evaluation metrics grounded in logical correctness, and comparative studies of different LLM architectures and training strategies. This work substantially extends \cite{Horner2025}. Notable improvements include a revised formula for the success score that more precisely reflects the extent to which all aspects of a law text have been formalized by a LLM, evaluations of newly released models and the introduction of a two-stage pipeline with a refinement step. Additionally, the evaluation procedure has been refined through stricter error assessment.

The remainder of this paper is structured as follows.
Section~\ref{sec:background} provides the necessary background on LLMs and the formal representation language DDL used in this study.
Section~\ref{sec:methodology} outlines the methodology, including the segmentation of legal texts into individual law snippets, their transformation into DDL, and the evaluation approach.
Section~\ref{sec:results} presents the experimental results, covering prompt engineering, multi-snippet processing strategies, fine-tuning, and two-stage pipelines. This section also includes a detailed comparison with the evaluation framework proposed by \cite{DBLP:conf/aicol/DragoniVRG17}.
Section~\ref{sec:limitations} discusses key limitations, such as challenges in legal implementation and handling of inter-snippet references.
Finally, Section~\ref{sec:conclusion-and-future-work} concludes the paper and outlines directions for future research.

\section{Background}
\label{sec:background}

\textbf{Defeasible Deontic Logic} \citep{jpl:permission,handbook:deontic} is a flexible and efficient rule-based non-monotonic formalism for the representation of legal norms and legal reasoning. The logic combines features of Defeasible Logic for the natural modeling of exceptions and defeasibility with concepts from Deontic Logic (i.e., obligations, permission, prohibition, compensatory obligations).  A rule in DDL has the form 
\[
    r\colon a_1, \dots, a_n \Rightarrow c
\]
where $r$ is the label (or name of the rule), $a_1,\dots,a_n$ are the premises of the rule, and $c$ is the conclusion of the rule. $a_1,\dots,a_n,c$ are either literals or deontic literals, where a literal is either an atomic proposition or its negation, and a deontic literal is a literal in the scope of a deontic operator ([O] for obligation, [F] for forbidden or prohibition, and [P] for permission).  Moreover, the logic is equipped with a superiority relation, a binary relation over the set of rules. The superiority relation is used when two conflicting rules are both applicable, and specifies which rule prevails over the other. 

The DDL reasoning mechanism has an argumentation-like structure. To prove a conclusion, we need to have an applicable rule for it. Then we have to consider all possible counterarguments, namely the rules for the opposite. For each of such rules, we have to rebut them. Thus, we have to either discard it (show that the rule is not applicable) or defeat it, which means we have to show an applicable rule that defeats it (using the superiority relation). 

{\bf Large Language Models (LLMs)} are advanced machine learning systems designed to understand and generate human language. Trained on vast amounts of textual data, LLMs are capable of performing a wide range of language-related tasks, including text generation, summarization, translation, question answering, and formal reasoning. They achieve this by learning complex statistical patterns and representations of language, enabling them to predict the most likely continuation of a given input.

LLMs can be broadly divided into two categories: \textit{traditional LLMs} and \textit{reasoning LLMs}.
Traditional LLMs, such as the GPT series by OpenAI, DeepSeek-V3 or similar models, are primarily optimized for fluent language generation and general-purpose tasks. Their strength lies in producing coherent and contextually appropriate text. However, their capabilities in logical reasoning and structured problem-solving are limited.
Reasoning LLMs represent a newer generation of models that are explicitly designed to perform structured reasoning tasks more effectively. These models consider additional training objectives, architectural innovations, or fine-tuning procedures that enhance their ability to perform logical inference, complex decision-making, and consistent multi-step problem-solving. Reasoning LLMs aim not only at linguistic fluency, but also at improved logical accuracy and reliability in formal contexts.

Recent advancements in LLMs offer opportunities to automatically extract formal semantics like DDL from legal documents.
However, achieving logical coherence and semantic validity remains non-trivial, motivating the need for careful experimental design.

In this study, we consider models from both categories: (i) Traditional LLMs: GPT-4.1, GPT-4o, GPT-4o mini, DeepSeek-V3 (both 0324 and the original version), Claude Sonnet 4 (without extended thinking), Llama 4 Maverick; and (ii) Reasoning LLMs: OpenAI o3, OpenAI o1, OpenAI o4-mini, OpenAI o3-mini, DeepSeek-R1 (both 0528 and the original version), Claude Sonnet 4 (with extended thinking).
Out of these considered models, all models of DeepSeek as well as Llama 4 Maverick are open-source.
These models will be evaluated and compared based on their performance in formalizing legal texts into DDL representations.

\section{Methodology}
\label{sec:methodology}
To reduce hallucinations and promote deterministic behavior, all LLMs were assessed under conservative decoding settings:

\begin{itemize}
    \item \texttt{temperature = 0}  \hfill
    
    The \texttt{temperature} parameter controls the degree of randomness in the generation process. 
    Higher values lead to more diverse outputs\mbox{~\citep{DBLP:conf/icccrea/PeeperkornK0J24}}, facilitating exploratory or creative behavior~\mbox{\citep{DBLP:conf/ccnlg/ManjavacasKBK17}}.
    Therefore, \texttt{temperature} was set to $0$.

    \item \texttt{top-p = 1}  \hfill
    
    Top-p sampling, also known as nucleus sampling, limits the pool of candidate tokens to those comprising the top cumulative probability $p$\mbox{~\citep{DBLP:conf/iclr/HoltzmanBDFC20}}. A value of $p=1$ effectively disables nucleus sampling, allowing all tokens to be considered. The parameter was set at its default value of $1$, following recommendations from both OpenAI and DeepSeek that advise adjusting either \texttt{temperature} or \texttt{top-p}, but not both simultaneously\mbox{~\citep{OpenAiApiReference, DeepSeekApiReference}}.

    \item \texttt{frequency penalty = 0} \hfill
    
    This parameter discourages the repeated generation of tokens that have already appeared by reducing their probability proportionally to their frequency\mbox{~\citep{DBLP:journals/corr/abs-2408-10577}}. 
    While beneficial for enhancing lexical diversity, it is unsuitable for the present task, where consistent use of terminology across similar legal contexts is required. Therefore, the parameter was left at its default setting of $0$.

    \item \texttt{presence penalty = 0} \hfill
    
    The \texttt{presence penalty} penalizes the generation of tokens that have already been produced at least once, independently of their frequency\mbox{~\citep{DBLP:journals/corr/abs-2408-10577}}. Its purpose is to encourage novel token introduction. However, similar to the frequency penalty, 
    consistent lexical choice is essential for reliable formalization, so this parameter was kept at its default value of $0$.
\end{itemize}

These settings were applied both during legal text segmentation and formalization into DDL. 
For OpenAI's reasoning models, which do not accept these parameters, the \texttt{reasoning\_effort} option was set to \texttt{high}. Also for Claude Sonnet 4 with extended thinking enabled, these options were not available.

The rest of this section presents the core aspects that form the basis of our approach.

\subsection{Segmentation into Law Snippets}
\label{method-splitting}
Legal texts are initially segmented into manageable ''law snippets`` using DeepSeek-R1.
A key challenge in instructing the LLMs was determining the optimal length for law snippets: overly long snippets risked losing critical information during formalization, whereas overly short ones hindered atom reuse. To address this, we instructed the model to split enumerations containing more than two elements into separate law snippets while preserving shorter ones intact.

Note that pre-processing might not be required for all legal texts. In some normative acts, paragraphs are sufficiently short and need no further subdivision. However, in documents like the Australian Telecommunications Consumer Protections Code, where individual articles can span 4-5 pages, splitting the text into smaller segments helps the LLM systematically analyze each component without overlooking critical details.

\subsection{Transformation into DDL}
We evaluated three main strategies:

\begin{itemize}
    \item \textbf{Chain-of-Instructions (CoI)} prompting with varying configurations and shot counts.
    \item \textbf{Fine-tuning of GPT-4.1 and GPT-4o} using a limited dataset of annotated examples to enhance task-specific performance.
    \item \textbf{Two-stage pipelines}, where we explored two approaches: (i) using a separate LLM for atom extraction, and (ii) introducing a refinement step after the initial generation.
\end{itemize}

Despite adhering to the guidelines for achieving reproducible outputs by \cite{OpenAiReproducibleOutputsReference}, e.g., fixing the \texttt{seed} and \texttt{tem\-per\-a\-ture} parameters (when supported by the model), we encountered non-deterministic behavior. 
This phenomenon can be attributed to inherent LLM stochasticity~\citep{blackwell2024}. 
Note that this behavior was also observed in the open-source models considered in this study.

\subsection{Evaluation}

We evaluate the generated rules across six dimensions, each operationalized as a concrete question: completeness (Q1), syntactic (Q2) and semantic correctness (Q3), deontic modality accuracy (Q4), precondition appropriateness (Q5), and meaningfulness/reuse of atom names (Q6).

It is important to note that a single law snippet may lead to the generation of multiple rules. Furthermore, Q1 is assessed based on the law snippet as a whole, taking into account all rules derived from it.
In contrast, Q2 through Q6 are evaluated individually for each generated rule.
The questions are ordered such that earlier ones address more general and fundamental aspects of correctness, while later ones examine increasingly fine-grained details.
Importantly, the evaluation follows a short-circuiting scheme: if for some rule $r$ a question Qi with $i \geq 2$ is evaluated as \texttt{false}, then all subsequent questions Qj with $j > i$ are not considered and are implicitly assigned the value \texttt{false}.

\newcommand{\questionone}{Are all aspects of the law text formalized?}

\medskip\noindent\textbf{[Q1: Completeness.] \questionone} \\
Consider, for instance, the following formalization of law snippet 8.2.1.a.xiv, which states that a supplier may close a complaint only with the consumer's consent or in compliance with clauses 8.2.1(c), (d), or (e).

\begin{ruletable}
    \tablecenter{complaint(X), consentConsumer(X) $\Rightarrow$ [P] closeComplaint(X)} \\\addlinespace
    \tablecenter{complaint(X), complied8.2.1.c(X) $\Rightarrow$ [P] closeComplaint(X)} \\\addlinespace
    \tablecenter{complaint(X), complied8.2.1.d(X) $\Rightarrow$ [P] closeComplaint(X)} \\\addlinespace
    \tableleft{complaint(X), complied8.2.1.e(X) $\Rightarrow$ [P] closeComplaint(X)}
\end{ruletable}

This is not a complete formalization of the facts, as the following rule is missing:
\begin{ruletable}
    \tablecenter{complaint(X) $\Rightarrow$ [O] -closeComplaint(X)}
\end{ruletable}

This initial check is crucial to prevent the LLM from achieving a high score merely by formalizing only the simplest aspects of a problem. 
Unlike subsequent dimension evaluations, however, Q1 is evaluated once per law snippet, not once per generated rule. 
Another important distinction between Q1 and the other checks is that the answer in this case is not a binary \texttt{true}/\texttt{false} value. Instead, it is expressed as a percentage indicating the proportion of gold standard rules for the respective law snippet that were \emph{attempted} to be formalized.
A rule is considered attempted if it can be meaningfully matched to a gold standard rule, even if the generated version contains minor flaws (e.g., suboptimal atom naming, redundant preconditions).
The key point is that Q1 measures attempted coverage (matching effort), not perfect correctness. The correctness of those attempted rules is then picked up progressively downstream in Q2--Q6.

\newcommand{\questiontwo}{Is the rule syntactically valid and non-redundant?}

\medskip\noindent\textbf{[Q2: Syntactic Validity.] \questiontwo} \\
An example of a rule that fails the syntactic validity check is the following:
\begin{ruletable}
    closeComplaint(X), -consent(X), -clausesCDEComplied(X) $\Rightarrow$ [O] closeComplaint(X)
\end{ruletable}

Note that the consequence of this rule also appears as its antecedent. 
Furthermore, this question checks for syntactically identical rules that appear more than once within a law snippet.

\newcommand{\questionthree}{Is the rule semantically valid and non-redundant?}

\medskip\noindent\textbf{[Q3: Semantic Correctness.] \questionthree} \\
This question serves as a ``catch-all'' check that applies when no other question describes the problem better, for example, when hallucinations of the LLMs occur. The following rules fail this check, as the atoms informResolution(X) and informNoResolution(X) are unrelated to the facts described in the legal text.
\begin{ruletable}
    \tablecenter{informResolution(X) $\Rightarrow$ [P] closeComplaint(X)} \\\addlinespace
    \tablecenter{informNoResolution(X) $\Rightarrow$ [P] closeComplaint(X)}
\end{ruletable}

However, there are also more subtle issues filtered by this question, for example, when a LLM combines several aspects with a logical ``and'', even though they should be connected with a logical ``or'' according to the legal text. This question also verifies whether two rules of a formalization convey the same meaning and are therefore redundant, even though they are not syntactically identical.

\newcommand{\questionfour}{Are the Deontic modalities and negations correctly placed?}

\medskip\noindent\textbf{[Q4: Deontic Modality Accurracy.] \questionfour} \\
In this example, a permission is incorrectly formalized as an obligation:
\begin{ruletable}
    \tablecenter{complaint(X), consentConsumer(X) $\Rightarrow$ [O] closeComplaint(X)}
\end{ruletable}

Hence, the question would be answered with \texttt{false}, and no further checks performed. Note that this output stems from an early variation of the prompt. 
Such an error did not occur in later iterations.

\newcommand{\questionfive}{Is the precondition appropriate?}

\medskip\noindent\textbf{[Q5: Precondition Appropriateness.] \questionfive} \\
A common problem was that the precondition of the rules contained either too many, too few or wrong atoms. This question should cover precisely these cases.

Consider for instance the following formalization generated in an experiment:

\begin{ruletable}
    \tablecenter{consentConsumer(X) $\Rightarrow$ [P] closeComplaint(X)} \\
    \addlinespace
    \tablecenter{compliedWithClauseC(X) $\Rightarrow$ [P] closeComplaint(X)} \\\addlinespace
    \tablecenter{compliedWithClauseD(X) $\Rightarrow$ [P] closeComplaint(X)} \\\addlinespace
    \tablecenter{compliedWithClauseE(X) $\Rightarrow$ [P] closeComplaint(X)} \\\addlinespace
    \tableleft{-consentConsumer(X), -compliedWithClauseC(X),-compliedWithClauseD(X),} \\
    \tableright{-compliedWithClauseE(X) $\Rightarrow$ [F] closeComplaint(X)}
\end{ruletable}

In the last rule, it is not necessary that all these atoms are included in the precondition. A simple complaint(X) would have been enough -- that the prohibition to close the complaint does not hold when there is consent from the consumer already follows from the first rule.

\newcommand{\questionsix}{Are the atom names meaningful and, if appropriate, reused?}

\medskip\noindent\textbf{[Q6: Meaningfulness/Reuse of Atom Names.] \questionsix}\\
Consider again the above formalization, for example, the atom compliedWithClauseC(X). Unfortunately, it is not fully clear from the atom name to which clause the name is referring -- a better name would be \text{clause8.2.1cComplied(X)}.

\medskip
Let $l$ be a law snippet. We denote by $\mathcal{R}(l)$ the set of rules generated by the LLM from $l$ and by $\mathcal{G}(l)$ the set of rules from the gold standard corresponding to $l$.
We write $r \sim r'$ if a rule $r \in \mathcal{R}(l)$ can be meaningfully matched to a rule $r' \in \mathcal{G}(l)$ despite possible minor errors as explained earlier.
To assess the quality of the generated formalization $\mathcal{R}(l)$, we define a success score $S(l) \in [0,1]$ that captures both the coverage of $\mathcal{G}(l)$ and the correctness of the generated rules in $\mathcal{R}(l)$.

Let $\text{Q1}(l) \in [0,1]$ denote the proportion of rules in $\mathcal{R}(l)$ w.r.t. $\mathcal{G}(l)$ that were \emph{attempted} to be formalized by the LLM.
For instance, if the gold standard snippet contains two rules and the LLM generates two flawed but still alignable rules, we assign $\text{Q1}(l) = 1.0$.
Conversely, if the LLM generates only one rule that matches one of the two gold standard rules, we assign $\text{Q1}(l) = 0.5$.
In this way, Q1 enforces that all gold standard rules should be \emph{addressed}, even if not perfectly realized.
Formally, $\text{Q1}(l)$ can be defined as:
\[
\text{Q1}(l) \;=\; \frac{\bigl|\{\, r \in \mathcal{R}(l) \;\mid\; \exists\, r' \in \mathcal{G}(l) \text{ such that } r \sim r' \}\bigr|}{|\mathcal{G}(l)|}.
\]

For each rule $r \in R(l)$, we then evaluate the remaining five binary criteria $\text{Q2}(r), \ldots, \text{Q6}(r) \in \{0,1\}$, asked in sequential order and designed to assess progressively finer aspects of correctness.
If any $\text{Qi}(r)$ evaluates to $0$, all subsequent questions for that rule are also set to $0$.

The {\bf success score $S(l)$} for an individual law snippet $l$ is then defined as:
\begin{equation*}
    S(l) = \text{Q1}(l) \times \frac{1}{|\mathcal{R}(l)|} \sum_{r \in \mathcal{R}(l)}  \frac{1}{5} \sum_{i=2}^{6} \text{Qi}(r).
\end{equation*}
This formulation ensures that Q1 rewards coverage (whether an LLM attempts to capture all gold standard rules), while Q2--Q6 reward the accuracy of those attempts.

The {\bf overall success score $S(\mathcal{L})$} for a set of law snippets $\mathcal{L}$ is the average success score of the individual snippets:
\begin{equation*}
    S(\mathcal{L}) = \frac{1}{|\mathcal{L}|} \sum_{l \in \mathcal{L}} S(l).
\end{equation*}
In addition, we define a stricter evaluation $S^{*}$ where only perfect formalizations contribute to the success score:
\[
S^{*}(l) = 
     \begin{cases}
      1 & \text{if } S(l) = 1, \\
      0  & \text{otherwise}
    \end{cases}, \quad
S^{*}(\mathcal{L}) = \frac{1}{|\mathcal{L}|} \sum_{l \in \mathcal{L}} S^{*}(l).
\]

\section{Results}
\label{sec:results}

A series of experiments has been conducted to identify the LLM configuration that is the most promising for the formalization task. 
These experiments are conducted on real legal content from Sections 8.2.1(a)–(c) of the TCP Code.

In each of the following experiments, 
we provided LLMs with a prompt containing step-by-step instructions to guide the model through its task.
This approach is called Chain-of-Instructions (CoI) prompting \citep{DBLP:conf/jurix/ZinSB24}.
Hence, the model is encouraged to solve each subtask step by step until the final answer is reached \citep{DBLP:journals/corr/abs-2402-11532}.
This method contrasts with Chain-of-Thoughts (CoT), which usually depends more on implicit reasoning \citep{DBLP:conf/jurix/ZinSB24} -- especially for Zero-Shot-CoT, where just a sentence like \textit{``Let's think step by step''} is appended to the prompt \citep{DBLP:journals/corr/abs-2401-14295}.

Moreover, we use few-shot learning~\citep{DBLP:conf/nips/BrownMRSKDNSSAA20}, where we provide the LLM with a few examples (input-output pairs) in the prompt to demonstrate 
how to solve the task.

In all experiments conducted, the prompt was passed to the LLMs via a \texttt{system} message.

\subsection{Prompt Development}
\label{determining-the-best-prompt}
The prompt employed in our experiments was derived through a series of iterative refinements.
Beginning with an initial two-shot learning prompt, successive modifications were made to enhance the clarity of the instructions and the quality of the generated outputs.
These iterations involved the inclusion of additional guidance and examples to better align the model's behavior with the desired output format.
The final version utilizes a three-shot learning approach.
Listing \ref{lst:prompt-best} provides the complete and final prompt used in our evaluation.

\begin{lstlisting}[frame=single,breaklines,basicstyle=\ttfamily\footnotesize,captionpos=b,caption={Best prompt},label={lst:prompt-best}]
Transform legal text in natural language to expressions in Defeasible Deontic Logic 
(DDL) in XML format. Each  atom should end with "(X)". If you want to represent a 
conjunction, separate the atoms by a comma. If you want to represent a disjunction, 
please use multiple rules and do not write it as a single rule. Output only a  
single <Paragraph> element with multiple <Rule> elements if necessary. Make sure to
output valid XML. Represent obligations with [O], permissions with [P], and  
prohibitions with [F]. If you want to negate an atom, use the negation symbol "-" 
before the atom or the deontic operator. Each rule should have only one
consequence. If you want to represent multiple consequences, please use multiple 
rules. Since the law snippet are talking about complaint handling, in most 
preconditions, there will be an atom like complaint(X). Make sure to keep the atoms
in the precondition as simple as possible. If it is possible to break down the 
atoms into smaller parts, please do so. For example, instead of urgentComplaint(X),
write complaint(X), urgent(X). Moreover, NEVER put an atom in the antecedent if it
also appears in the consequence, because this would be syntactically invalid.

Work in the following steps:
1. Define the atoms that will be used in the rules.
2. Define the if-then structure of the rules.
3. Identify deontic modalities.
4. Formalize the rules in the given format using  Defeasible Deontic Logic (DDL).

# Example 1
## Input
8.1.1 A Supplier must take the following actions to enable this outcome:
  (c) Ensure awareness and visibility: ensure their staff who have direct contact
  with Consumers or former Customers, including personnel working for contractors, 
  understand the Supplier's Complaint handling process, their responsibilities 
  under it and are able to identify and record a Complaint.
## Output
<Paragraph paragraphLabel="8.1.1.c">
  <Rules>
    <Rule ruleLabel="tcpc.8.1.1.c.1">
      complaintHandlingProcess(X) => 
        [O] relevantStaffAwareComplaintHandlingProcess(X)
    </Rule>
    <Rule ruleLabel="tcpc.8.1.1.c.2">
      complaintHandlingProcess(X) => [O] relevantStaffAbleToHandleComplaint(X)
    </Rule>
  </Rules>
</Paragraph>

# Example 2
## Input
8.1.1 A Supplier must take the following actions to  enable this outcome:
  (a) Implement a process: implement, operate and comply with a Complaint handling 
  process that:
    (x) is transparent, including:
      D. requiring Consumers or former Customers to be advised of the Resolution of
      their Complaint; and
## Output
<Paragraph paragraphLabel="8.1.1.a.x.D">
  <Rules>
    <Rule ruleLabel="tcpc.8.1.1.a.x.D">
      complaint(X), resolution(X) => [O] informResolution(X)
    </Rule>
  </Rules>
</Paragraph>

# Example 3
## Input
8.5.1 A Supplier must take the following actions to enable this outcome:
  (e) Maintain confidentiality: Suppliers not subject to the requirements of the 
  Privacy Act must ensure personal information concerning a Complaint is not 
  disclosed except as required to manage a Complaint with the TIO or with the 
  express consent of the Consumer.
## Output
<Paragraph paragraphLabel="8.5.1.e">
  <Rules>
    <Rule ruleLabel="tcpc.8.5.1.e.1">
      complaintHandlingProcess(X), personalInformation(X), -subjectPrivacyAct(X) =>
        [O] -discloseInformation(X)
    </Rule>
    <Rule ruleLabel="tcpc.8.5.1.e.2">
      personalInformation(X), requestFromTIO(X) => [O] discloseInformation(X)
    </Rule>
    <Rule ruleLabel="tcpc.8.5.1.e.3">
      consentDisclosurePersonalInformation(X) => [P] discloseInformation(X)
    </Rule>
  </Rules>
</Paragraph>
\end{lstlisting}

\begin{figure}[h]
    \centering
    \includegraphics[width=\imgwidth]{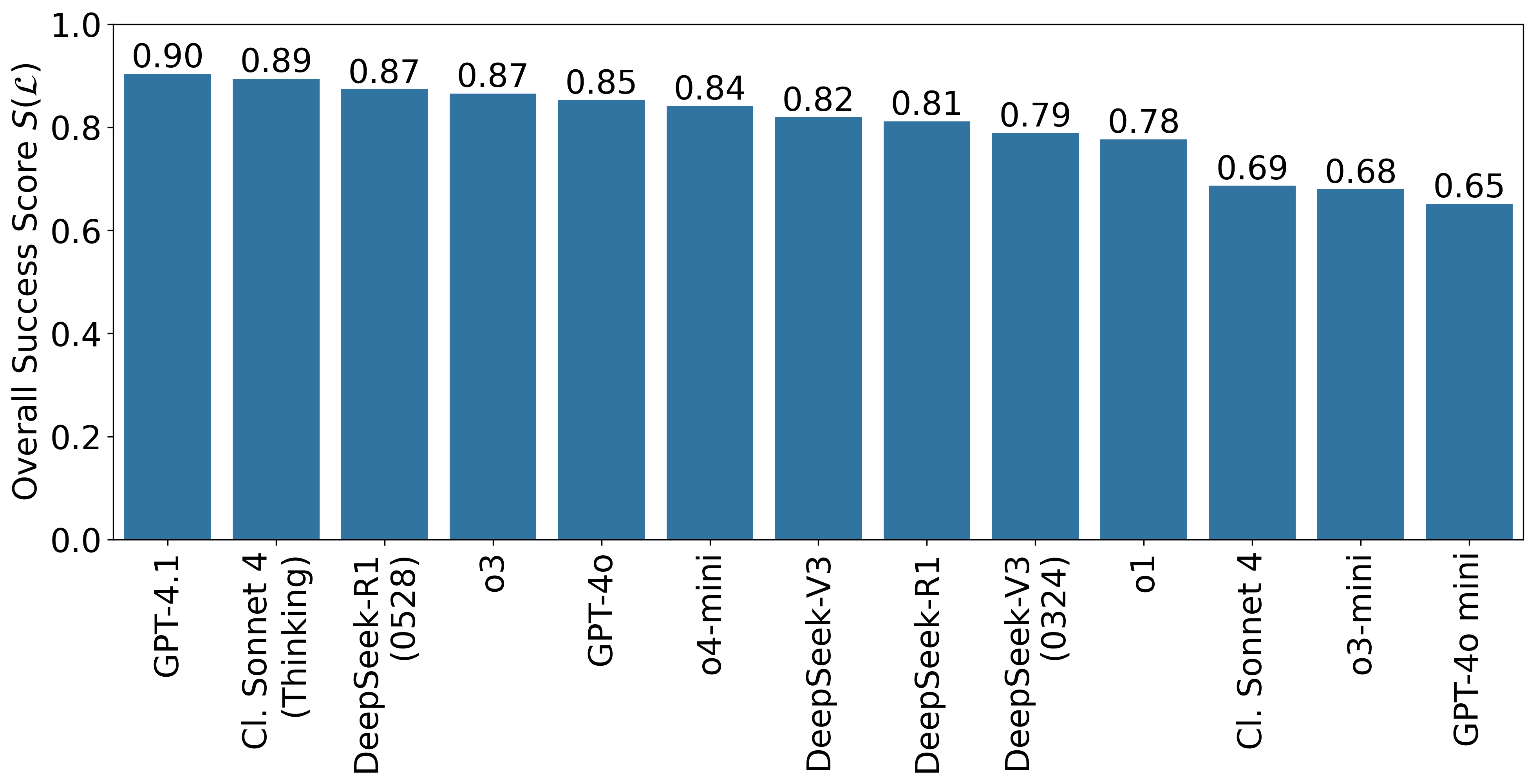}
    \caption{Success scores of various LLMs}
    \label{fig:all-llms}
\end{figure}

\begin{figure}[h]
    \centering
    \includegraphics[width=\imgwidth]{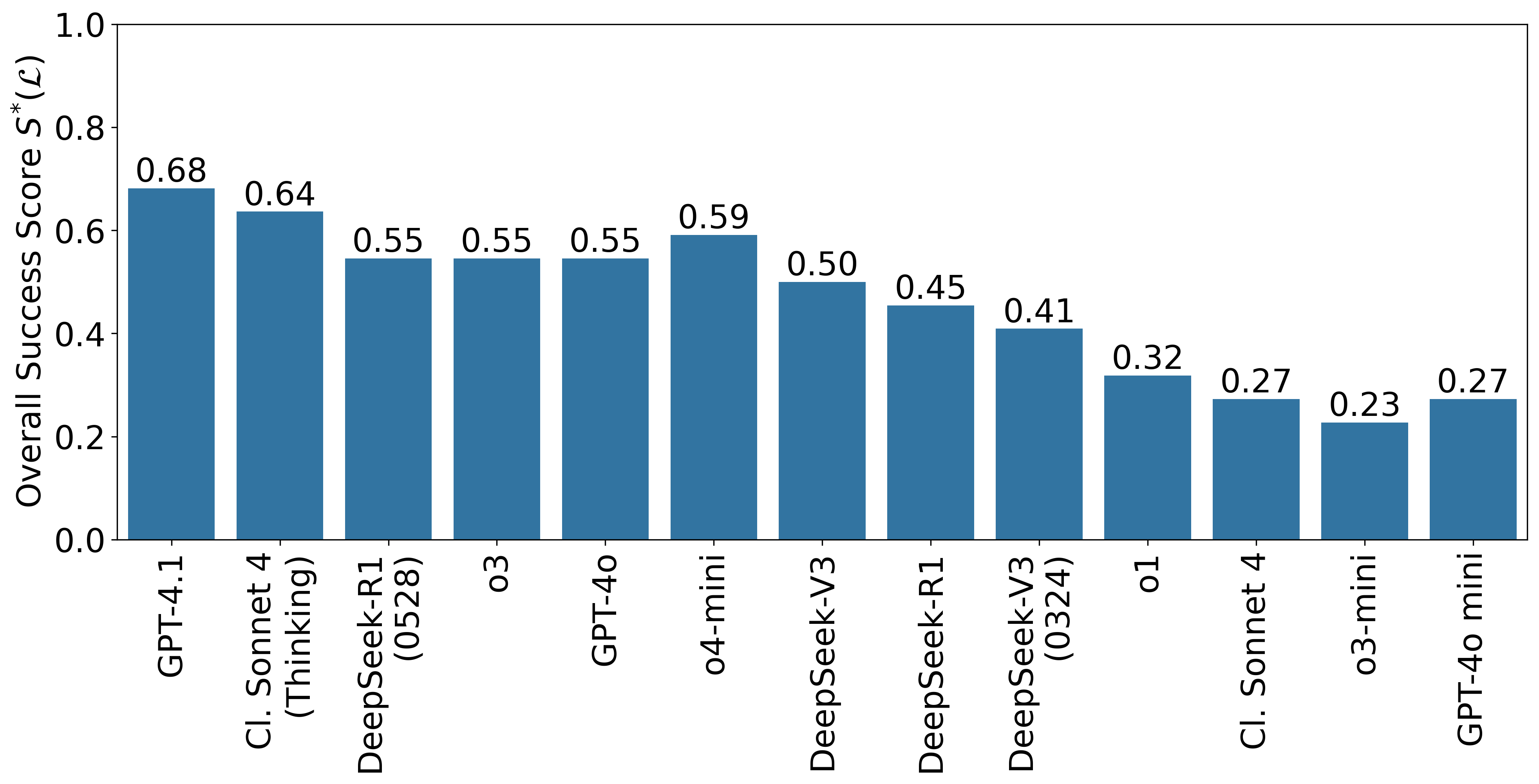}
    \caption{Success scores of all LLMs (perfect formalizations only)}
    \label{fig:all-llms-perfect}
\end{figure}

The final prompt was evaluated across multiple LLMs. 
Two diagrams summarize the results: one displaying the standard success scores $S$ (s. Figure~\ref{fig:all-llms}) and another illustrating the success scores $S^*$ under the stricter criterion of perfect formalizations (s. Figure~\ref{fig:all-llms-perfect}). An important observation is that the newly evaluated models generally outperform those considered in \cite{Horner2025}. Specifically, GPT-4.1 yields better results than GPT-4o, and DeepSeek-R1 (0528) surpasses the earlier release of DeepSeek-R1. This trend, however, does not extend to DeepSeek-V3. Moreover, Claude Sonnet 4 demonstrates excellent performance when the Extended Thinking feature is enabled, whereas the same model without this feature yields notably poor results.


\subsection{Consideration of Multiple Law Snippets Simultaneously}
\label{consideration-of-multiple-law-snippets}
In the experiments described in Section~\ref{determining-the-best-prompt}, the prompt was sent together with an individual law snippet to the LLM.
This approach ensured optimal focus of the LLM on all details of the individual law snippets but limited the reuse of atom names across snippets.
Here, we investigate whether providing multiple law snippets simultaneously enhances formalization performance.

\subsubsection{Incorporating the Formalization History}
\label{considering-the-whole-chat-history}
In one variant, the complete formalization history was included by alternating \texttt{user} and \texttt{assistant} messages for all prior snippets, aiming to encourage more consistent reuse of atom names across different law texts.

However, no improvement was observed compared to the single-snippet baseline; in fact, the success scores were marginally lower.
A plausible explanation is that the additional context overwhelmed the models, hindering their ability to focus effectively on the current snippet.
In particular, the goal of improving the reuse of atoms across multiple law snippets was not achieved.

\subsubsection{Providing Only Previously Formalized Atoms}
\label{passing-previously-formalized-atoms}
In a second variant, previously extracted atom names were provided collectively in a single \texttt{user} message, rather than replicating the entire prior dialogue history.
For each new law snippet, three messages were sent to the LLM: 
(1) a \texttt{system} prompt,
(2) a \texttt{user} message listing previously formalized atom names (cf. Listing~\ref{lst:previously-returned-atom-names}), and
(3) a \texttt{user} message containing the new law snippet to be formalized.

\begin{lstlisting}[frame=single,breaklines,basicstyle=\ttfamily\footnotesize,captionpos=b,caption={\texttt{user} message including previous atom names},label={lst:previously-returned-atom-names}]
Try to reuse the following atoms you have used for the formalization of previous paragraphs:
* complaint(X)
* madeInPerson(X)
* acknowledgeImmediately(X)
...
\end{lstlisting}

Although an increased reuse of atom names was observed, the atoms were often applied in inappropriate or irrelevant contexts.
As a result, this approach led to a greater number of hallucinations rather than an improvement in the formalizations.
Consequently, no further evaluation of this strategy was pursued.

\subsubsection{Formalizing All Law Snippets in a Single Interaction}
\label{single-interaction}

In a final approach, all law snippets were provided together within a single \texttt{user} prompt to the LLM.
While the input text contained multiple snippets, the division into distinct law snippets was preserved to encourage the model to treat each snippet individually.

\begin{figure}[h]
    \centering
    \includegraphics[width=\imgwidth]{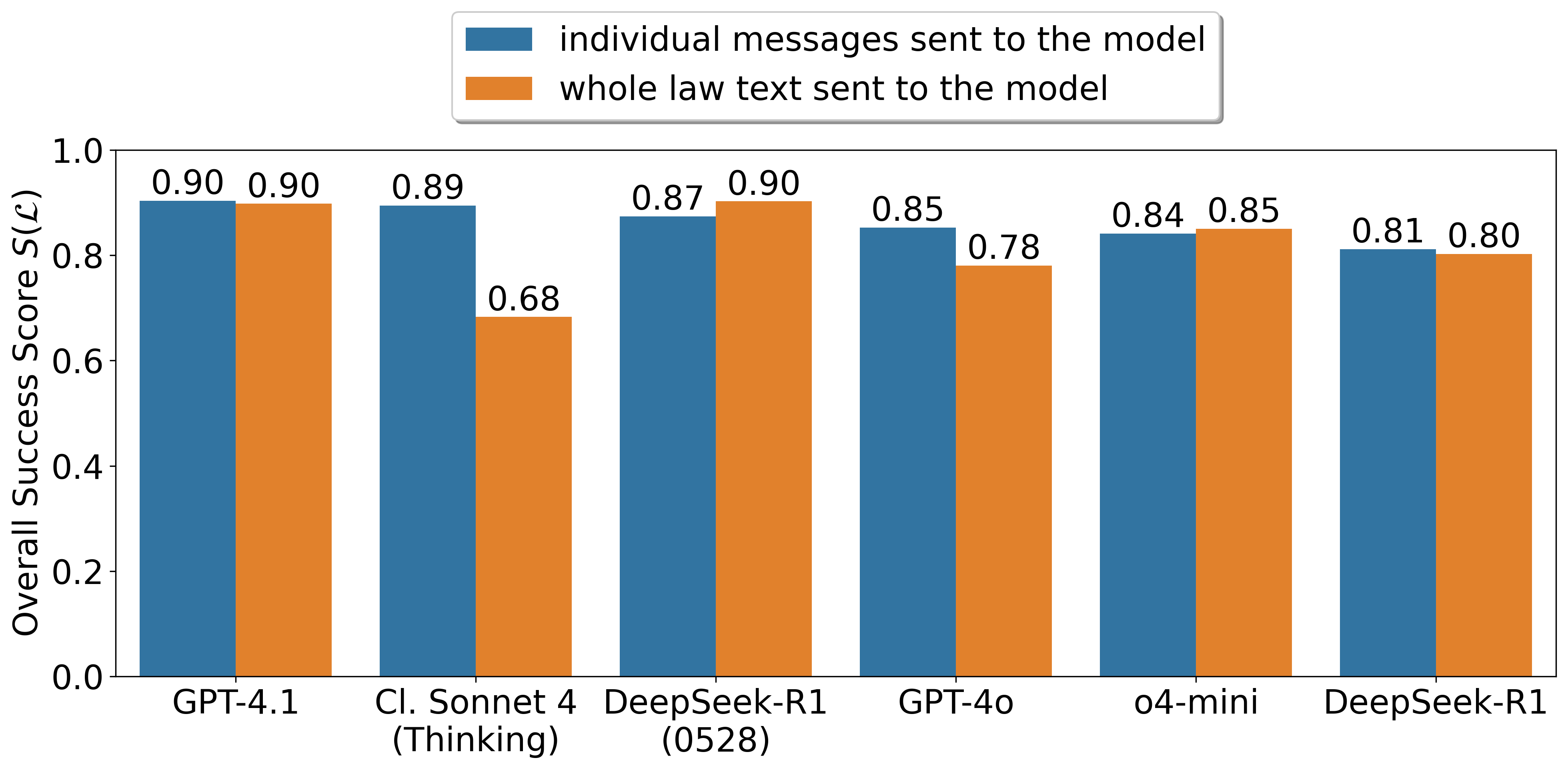}
    \caption{Success scores when formalizing all law snippets at once}
    \label{fig:single-chat}
\end{figure}
\begin{figure}[h]
    \centering
    \includegraphics[width=\imgwidth]{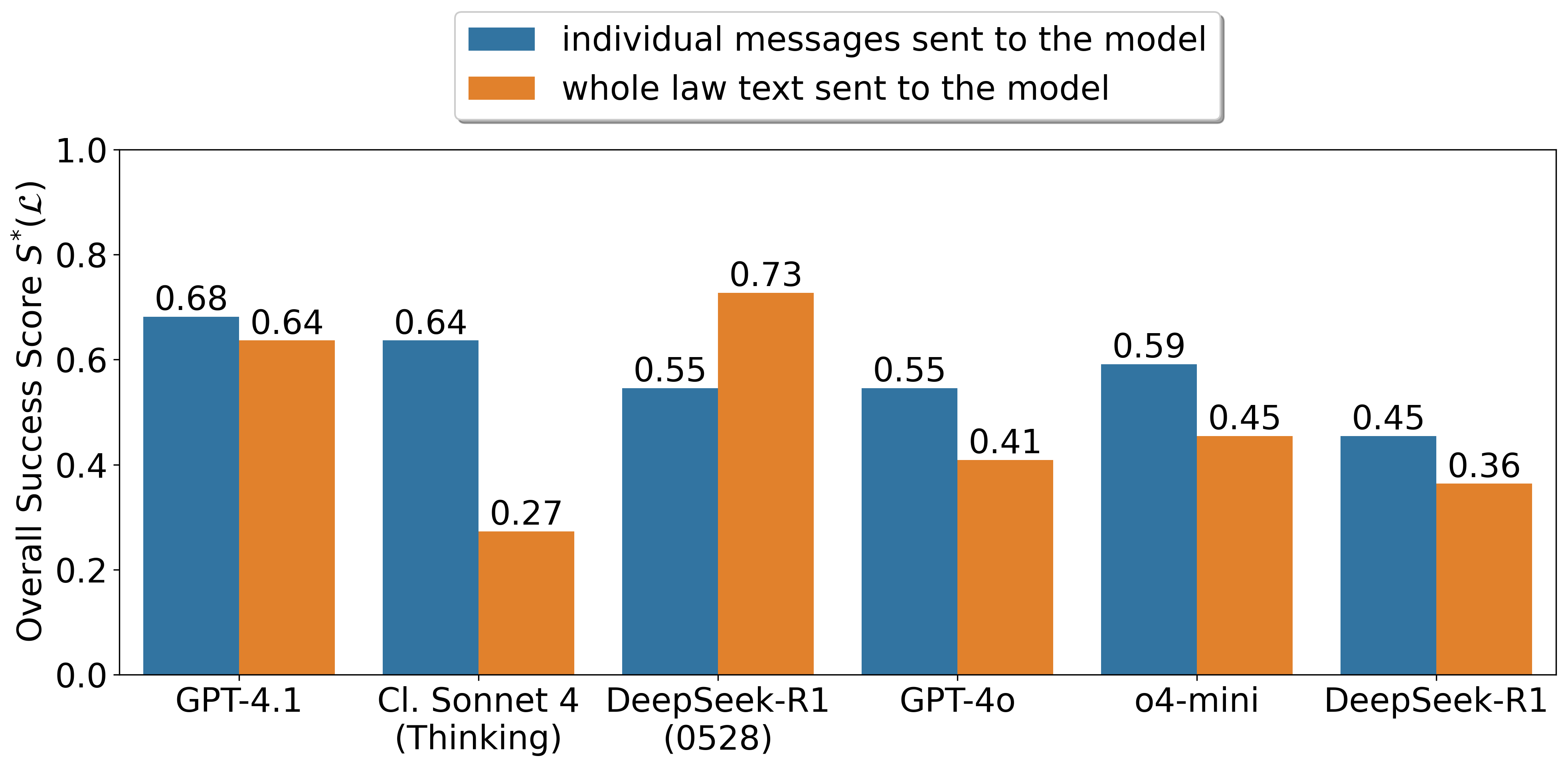}
    \caption{Success scores when formalizing all law snippets at once (perfect formalizations only)}
    \label{fig:single-chat-perfect}
\end{figure}

Note that it was not possible to evaluate OpenAI's o3 model in this experiment, as this model did not adhere to the predefined output structure and issued rules without further structuring in law snippets.

Figures~\ref{fig:single-chat} and~\ref{fig:single-chat-perfect} present the results for this setting, with the latter considering only perfect formalizations.

Although this approach led to a slight increase in atom reuse across snippets, it exhibited a major drawback: the generated formalizations were often less detailed compared to the baseline obtained in Section~\ref{determining-the-best-prompt}.
In particular, important facts were frequently merged into a single rule, even in cases where separate rules would have been necessary for a proper and precise formalization. This issue was especially pronounced for Claude Sonnet 4 with Extended Thinking enabled, though it was also observed in other models.

\subsection{Fine-Tuning}
\label{finetuning-the-models}

Fine-tuning is a transfer learning technique where pretrained model weights are adapted to a new task through further training. 
By leveraging knowledge acquired during pretraining, fine-tuning can substantially enhance model performance, particularly in scenarios characterized by limited training data.
Prior work has demonstrated the effectiveness of fine-tuning LLMs in improving task-specific outcomes~\citep{DBLP:conf/iclr/WeiBZGYLDDL22}.

In the present study, fine-tuning experiments were conducted with GPT-4o and GPT-4.1.
Due to the proprietary nature of OpenAI's GPT series, direct access to the model weights is not available. However, OpenAI offers fine-tuning capabilities for selected non-reasoning models through its platform. 

Given that only 22 law snippets from the gold standard presented in~\cite{DBLP:conf/aicol/DragoniVRG17} correspond to Sections 8.2.1(a)–8.2.1(c), the remaining 44 snippets from unrelated sections were utilized as training data.

Three distinct fine-tuning configurations were evaluated, as summarized in Table~\ref{tab:fine-tuning-hyperparameters}.

\begin{table}[h]
    \centering
    \caption{Fine-tuning hyperparameter configurations}
    \label{tab:fine-tuning-hyperparameters}
    \begin{tabular}{cccc}
    \toprule
        & Config. 1 & Config. 2 & Config. 3 \\\midrule
Epochs & $3$ & $3$ & $3$ \\
Batch Size & $1$ & $4$ & $4$ \\
LR Multiplier & $2$ & $1.5$ & $1$\\\bottomrule
    \end{tabular}
    
\end{table}

Configuration 1 parameters were determined automatically by OpenAI, as recommended for initial fine-tuning attempts~\citep{OpenAiFineTuningReference}. However, early signs of overfitting motivated adjustments such as increasing the batch size and reducing the learning rate in Configurations 2 and 3. 

\begin{figure}[h]
    \centering
    \includegraphics[width=1\linewidth]{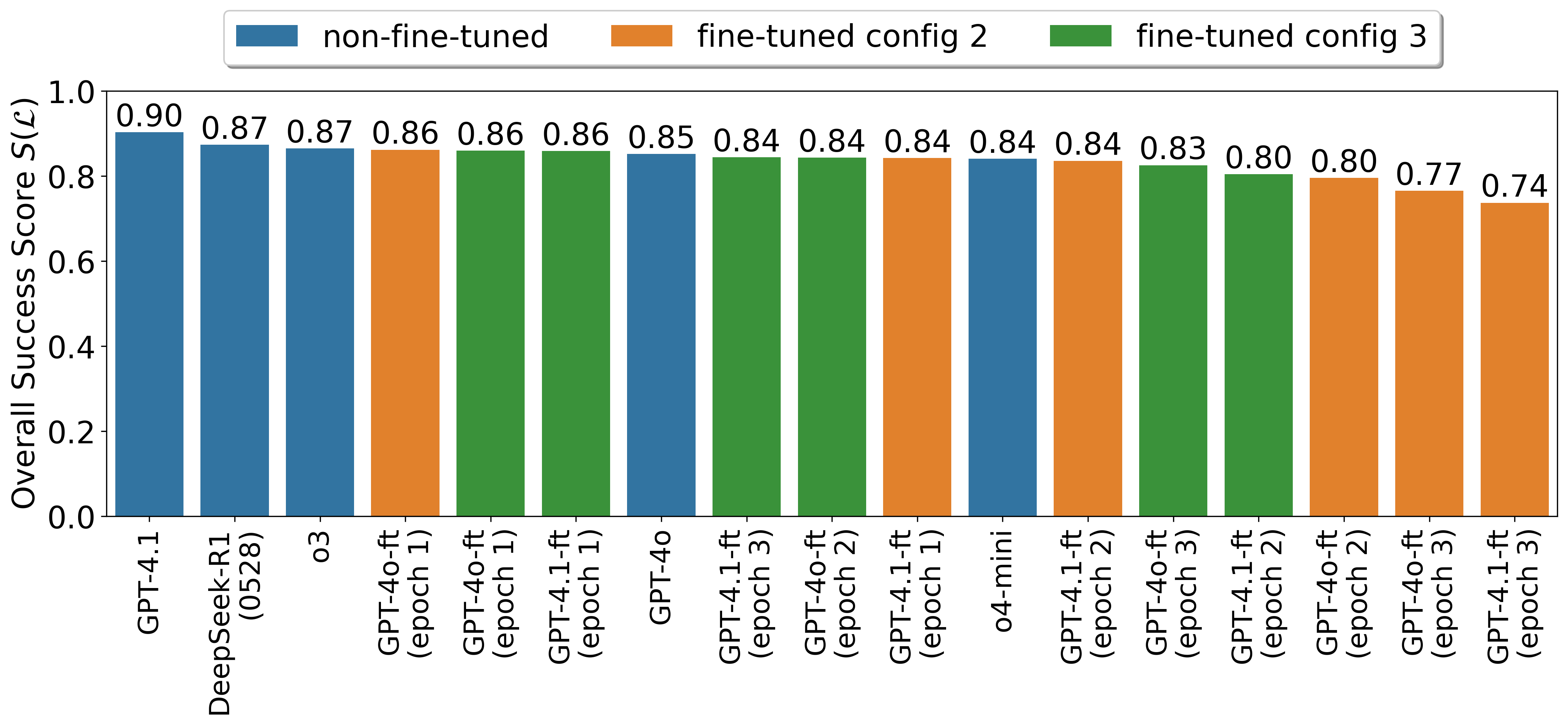}
    \caption{Success scores after fine-tuning}
    \label{fig:finetuned-model}
\end{figure}

\begin{figure}[h]
    \centering
    \includegraphics[width=1\linewidth]{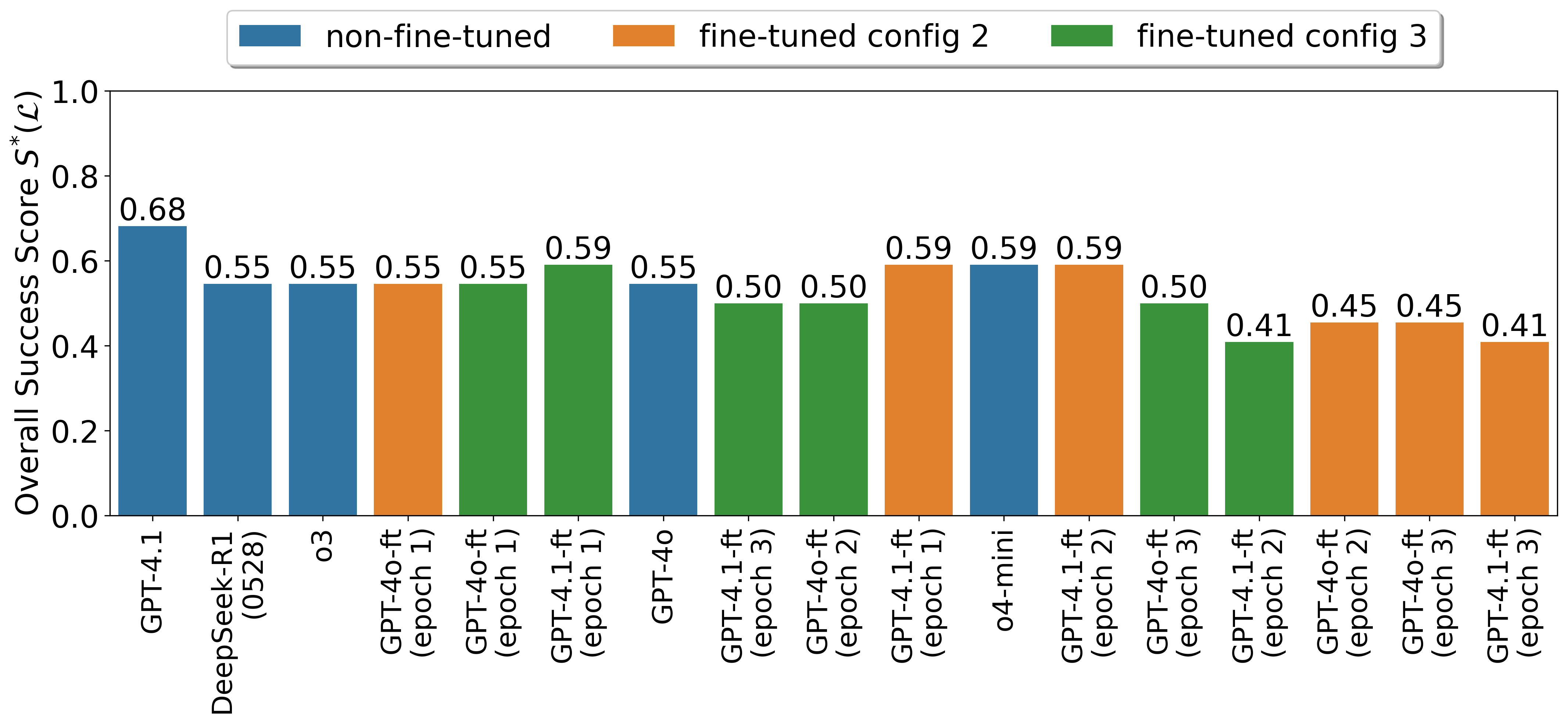}
    \caption{Success scores after fine-tuning (perfect formalizations only)}
    \label{fig:finetuned-model-perfect}
\end{figure}

After each training epoch, an evaluation has been conducted.
The resulting performance is depicted in Figures~\ref{fig:finetuned-model} and~\ref{fig:finetuned-model-perfect}, where blue bars correspond to non-fine-tuned baselines and other colors represent fine-tuned models. 
As the first configuration led to suboptimal results, its outcomes have been omitted from the figures for the sake of brevity and clarity.

For GPT-4o, fine-tuning resulted in a slightly improved success score after a single epoch of training under Configurations 2 and 3, relative to the baseline performance of the non-fine-tuned GPT-4o. However, subsequent training epochs led to a decline in performance, indicative of overfitting. Moreover, for GPT-4.1, fine-tuning led to less favorable results compared to using the base model without fine-tuning.


\subsection{Two-Stage Pipelines}
\label{implementing-a-multi-stage-pipeline}

In this approach, two different two-stage prompting strategies were employed, where the output of the first stage served as part of the input for the second stage.
This method aligns with the Layer-Of-Thoughts paradigm, which has been shown to enable complex reasoning in LLMs~\citep{DBLP:journals/corr/abs-2410-12153}.

\subsubsection{Separate Atom Extraction}
\label{separate-atom-extraction}

In this experiment, the intended names of the atoms alongside brief textual descriptions were extracted in the first stage from the legal texts using the system prompt in Listing~\ref{lst:prompt-atom-extraction}.
Three illustrative examples were provided within the prompt, thus applying a three-shot learning strategy.
This was followed by the formalization step based on the previously identified atom names.

\begin{lstlisting}[frame=single,breaklines,basicstyle=\ttfamily\footnotesize,captionpos=b,caption={Prompt for atom extraction (2 examples omitted)},label={lst:prompt-atom-extraction}]
Extract all the relevant atoms from the legal text in natural language and add a 
textual description of them. Each atom should end with "(X)". Do not include 
negations in the atom name - these will be introduced later on. Since the law 
snippet are talking about complaint handling, in most law snippets, there will be 
an atom like complaint(X). Make sure to keep the atoms as simple as possible. If 
it is possible to break down the atoms into smaller parts, please do so. For 
example, instead of urgentComplaint(X), write complaint(X), urgent(X). The only 
exception to this rule is when you can anticipate that an atom will belong into 
the consequence. In this case, a longer atom name is better, as each rule can have 
only one consequence. Keep in mind that these atoms will serve as antecedents and 
consequences in formalized rules - therefore, formalize enough atoms so that 
antecedents and consequents can be constructed from them. Formalize at least two
atoms.

# Example 1
## Input
8.1.1 A Supplier must take the following actions to enable this outcome:
  (a) Implement a process: implement, operate and comply with a Complaint handling
  process that:
    (v) clearly states that Consumers or former Customers have a right to make a 
    Complaint and that a proposed Resolution must be accepted by a Consumer or 
    former Customer before a Supplier is required to implement it;

## Output
informRightToMakeComplaint(X): Supplier informs customer of right to make a 
complaint.
informComplaintHandlingProcess(X): Supplier informs Customer of Complaint handling 
process.
complaintHandlingProcess(X): Supplier has a complaint handling process as per TCPC 
section 8.
\end{lstlisting}

For each model evaluated in Section~\ref{determining-the-best-prompt} (except Llama 4 Maverick and GPT-4o mini due to their poor previous performance), we conducted an experiment in this section for the atom extraction step. Moreover, a fine-tuned variant of GPT-4o, specifically trained on examples of atom extraction, was additionally examined.
As in Section~\ref{finetuning-the-models}, the 44 legal text snippets from~\cite{DBLP:conf/aicol/DragoniVRG17} not associated with Sections 8.2.1(a)–8.2.1(c) were used as training data for fine-tuning this model.

In the subsequent stage, the generation of DDL rules was performed based on the legal text and the previously extracted atom definitions. However, we observed that the overall performance remained relatively stable regardless of the model used in the second stage, which justifies limiting the evaluation at this stage of the pipeline to only a few selected models. Based on this observation, we conclude that the primary source of errors lies in the first step of the pipeline. Furthermore, we found that the initial models in the pipeline frequently generate superfluous atoms, which should not be considered by the model in the second step.

\begin{figure}[h]
    \centering
    \includegraphics[width=1\linewidth]{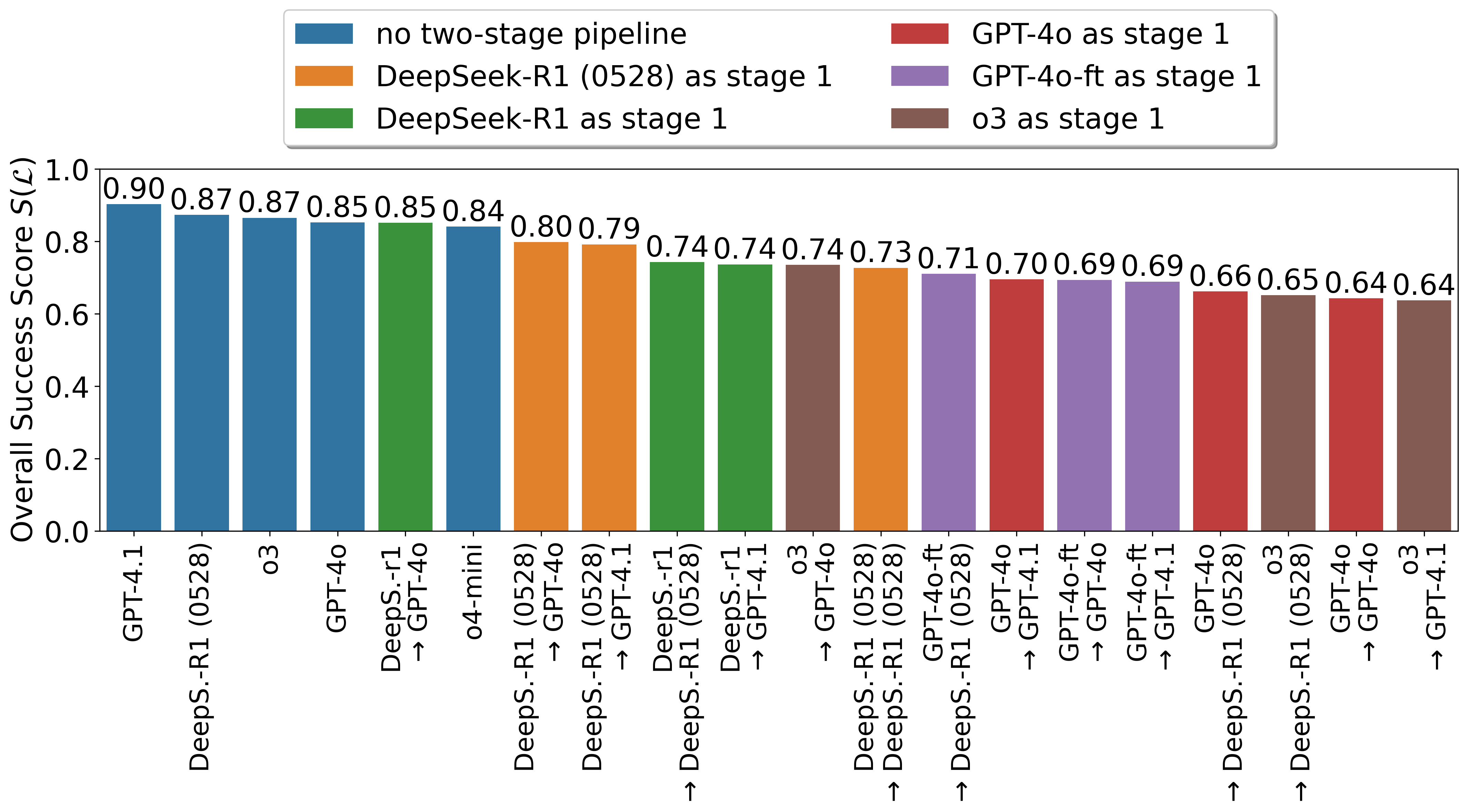}
    \caption{Success scores when using a separate atom extraction step}
    \label{fig:multi-step-pipeline}
\end{figure}

\begin{figure}[h]
    \centering
    \includegraphics[width=1\linewidth]{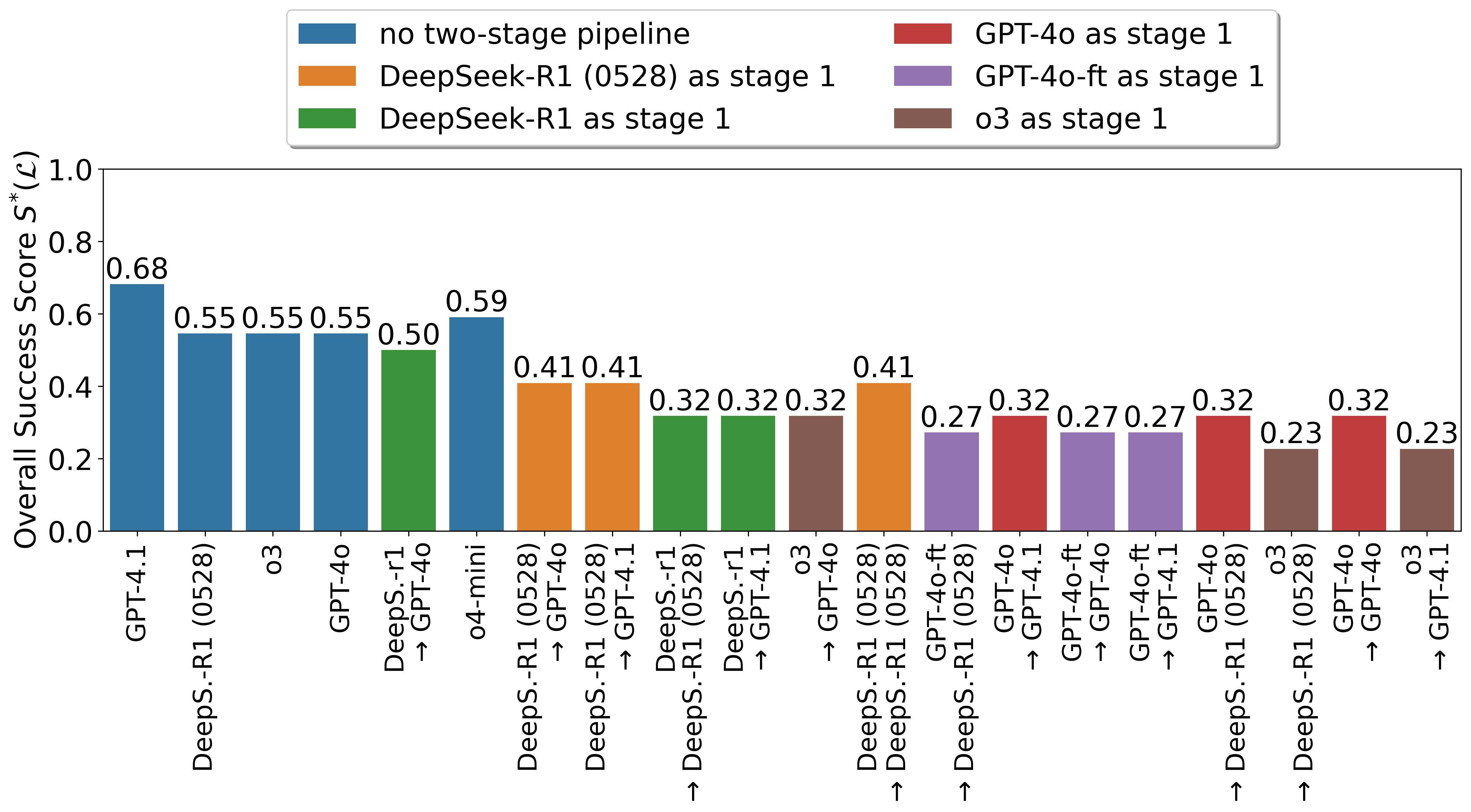}
    \caption{Success scores when using a separate atom extraction step (perfect formalizations only)}
    \label{fig:multi-step-pipeline-perfect}
\end{figure}

The results of these experiments are presented in Figures~\ref{fig:multi-step-pipeline} and~\ref{fig:multi-step-pipeline-perfect}.
Figure~\ref{fig:multi-step-pipeline} compares the success scores of the two-stage pipeline against those achieved with the best prompt from Section~\ref{determining-the-best-prompt} (blue bars).
Figure~\ref{fig:multi-step-pipeline-perfect} shows the corresponding comparison when only perfect formalizations are considered.


The figures indicate that this two-stage pipeline setup performs significantly worse than the new models presented in Section~\ref{determining-the-best-prompt}.

\subsubsection{Refinement Step}
\label{refinement-step}

We used the best formalizations from Section \ref{determining-the-best-prompt} as a basis for a refinement step using an additional LLM. 
The primary objective of this step was to mitigate systematic shortcomings observed in the earlier experiments, in particular the insufficient reuse of atom names across different law snippets.
This issue represents a substantial limitation of the prior results (including those discussed in Section~\ref{consideration-of-multiple-law-snippets}), as consistent atom reuse is essential for obtaining coherent and interpretable formalizations of legal texts.

For illustration, while the gold standard uniformly employs the atom \texttt{resolvable\-15Days(X)}, the LLM-generated formalizations frequently produced a wide range of variants such as \texttt{resolve\-In\-15Days(X)}, \texttt{resolved\-By15Days(X)}, or \texttt{cannot\-Resolve\-In15Days(X)}.
Such variation introduces unnecessary redundancy and hinders semantic consistency.
Moreover, it is undesirable for atom names themselves to contain lexical negations such as \enquote{not}, since negation is intended to be represented uniformly by a prefix operator (\texttt{-}) rather than encoded within the atom identifier.

In the refinement step proposed here, all law snippets, together with 
their corresponding generated rules, were processed jointly by Claude Sonnet~4.
The model was instructed to consolidate duplicate or inconsistent atom names into a unified vocabulary, thereby ensuring greater conceptual coherence across the formalizations.
The specific prompt used in this stage is provided in Listing~\ref{lst:prompt-refinement-step}.

\begin{lstlisting}[frame=single,breaklines,basicstyle=\ttfamily\footnotesize,captionpos=b,caption={Prompt for the refinement step},label={lst:prompt-refinement-step}]
In the following, I will give you an XML file which is composed of mulitple 
paragraphs, where each paragraph contains multiple generated rules. Please touch 
only the <Generated> element and its children. These generated rules formalize the
law text you find in the <RelevantText> atom using Defeasible Deontic Logic (DDL).
Regarding the syntax: A rule consists of multiple atoms as precondition, which are 
seperated by commas, and exactly one atom as conclusion. The conclusion is
separated from the preconditions by a "=>" sign (or =&gt;). Each atom name ends 
with (X). Atoms might be negated, which is indicated by a minus sign (-) in front 
of the atom name.

However, these formalizations are not perfect. Sometimes, there are (in different 
rules/paragraphs) multiple names introduced for the same concept. For example, 
"canResolveIn15Days(X)" and "resolvedIn15Days(X)" can be unified. Note that this 
is just an example. Please, look out for similar cases where two atoms mean the 
same thing and unify them. Pay attention! Look for atoms across ALL paragraphs/
rules that represent the same concept, not just within individual paragraphs. When
unifying atoms that represent the same core action but with different timing or
contextual constraints (for example adviseTimeframeWithin2Days(X) and advise-
TimeframeWithin15Days(X)), you may drop the timing constraint in order to re-use 
atoms more effectively. In general, the less atom names you have in total, the 
better.

Moreover, if it is not clear from the atom name what the atom is about, please make
the atom name more descriptive. For example, the following atom names are not well 
chosen:
* phoneCall(X) - it is not clear what happens in this phone call, better would be 
  "madeByPhoneCall(X)"
* unresolved(X) - contains a negation, which is not allowed
* acceptance(X) - better would be "customerAccepted(X)"
* complianceClauseC(X) - not clear what clause C is, better would be 
  "compliedWithClause8_2_1_c(X)"

If necessary, you may also remove atoms or add new ones, but make sure that the 
conclusion of the rules contains exactly one atom. Never add new <Rule> elements,
but only modify or delete existing ones.

Make sure that all atom names DO NOT contain negations like "not" or "cannot" - 
instead, write a minus sign (-) in front of the atom to show a negation. For 
example, if the atom name is "notUrgent(X)", you would write "-urgent(X)" instead.
Note that such negation words can also appear in the middle of an atom name. This 
allows you to reuse atoms both in positive as well as in negative contexts. This 
should also apply to atom names that you introduce yourself. In other words, in 
your output, no <Rule> element should contain an atom name that contains "not" or 
"cannot" or something similar. Attention: Do not simply remove the negation word,
but toggle the minus in front of the atom.

When you are done, output the full input I have given to you, but with the 
modifications incorporated that you have made. Do not give additional text or 
explanations, just the modified XML file.
\end{lstlisting}

Additional experiments were conducted using alternative models in the refinement step, including OpenAI's flagship models (o3, GPT-4.1, GPT-4o) as well as DeepSeek-R1.
However, these models yielded unsatisfactory results.
For instance, some failed to follow the instruction to output the complete XML representation, instead returning only partial fragments of the input.
Others generated atom names of excessive length, often extending to 100-150 characters, which rendered the output impractical for use.

\begin{figure}[h]
    \centering
    \includegraphics[width=\imgwidth]{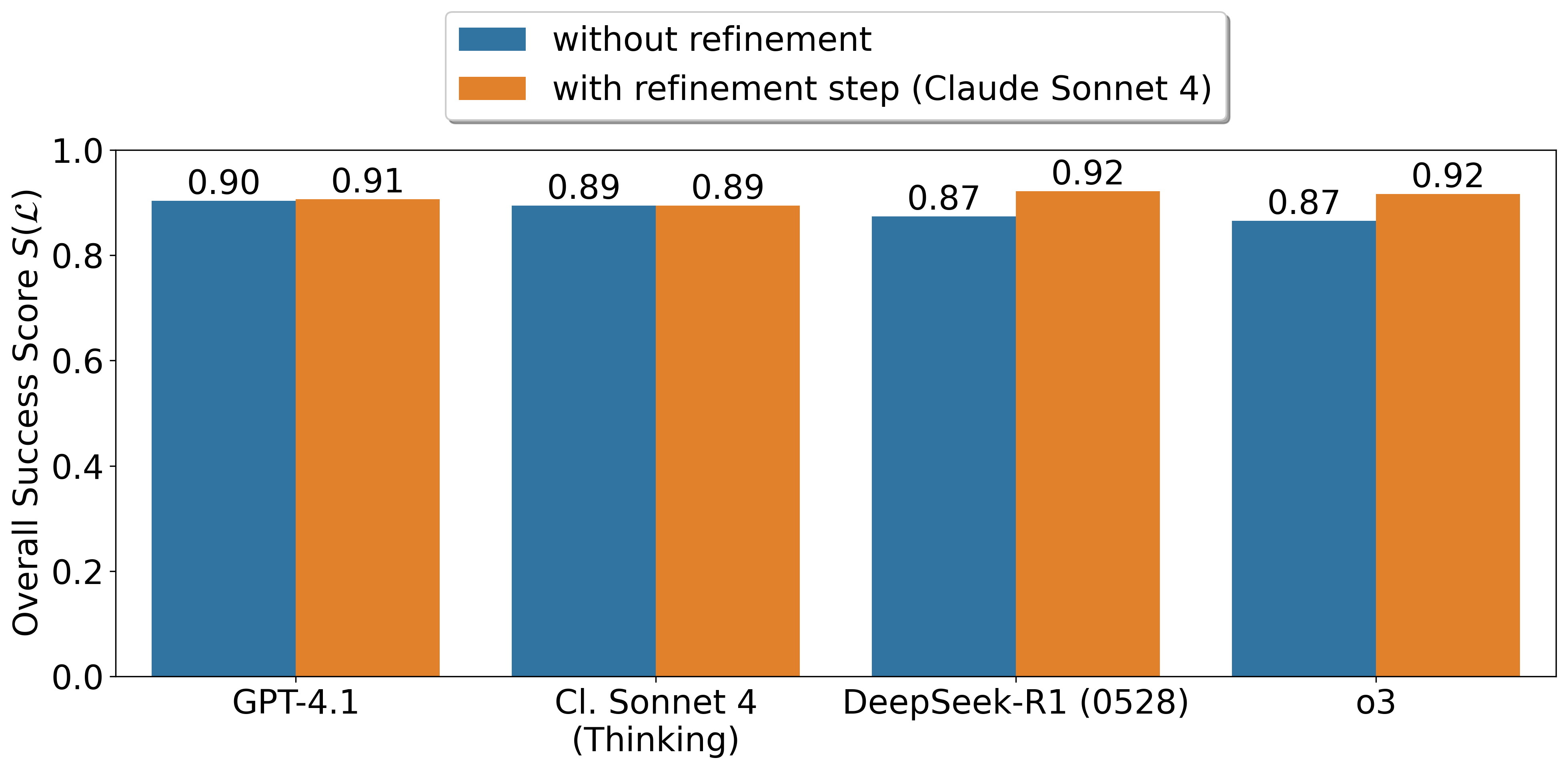}
    \caption{Success scores when using a refinement step}
    \label{fig:refinement-pipeline}
\end{figure}

\begin{figure}[h]
    \centering
    \includegraphics[width=\imgwidth]{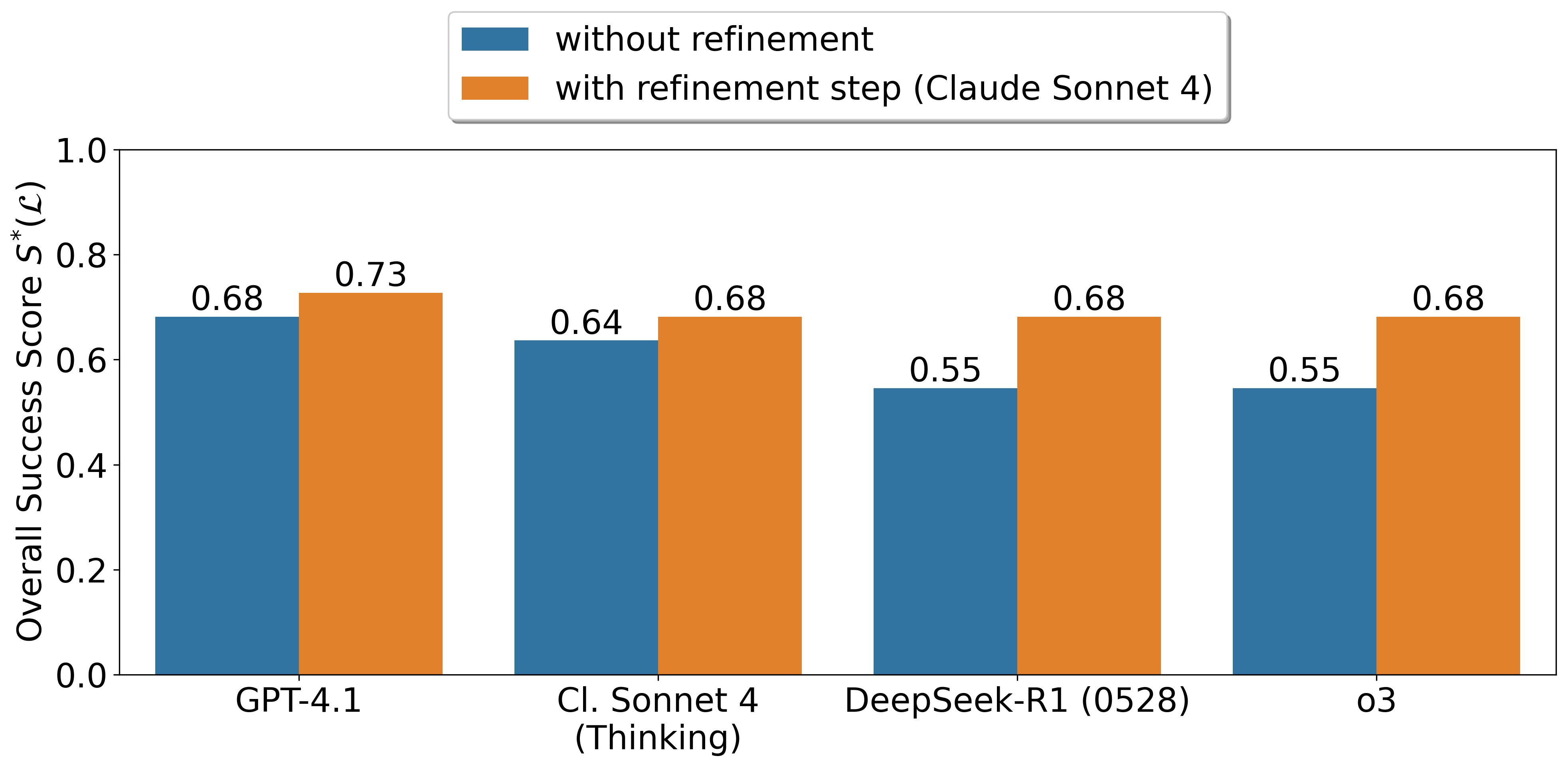}
    \caption{Success scores when using a refinement step (perfect formalizations only)}
    \label{fig:refinement-pipeline-perfect}
\end{figure}

The results of this refinement step are presented in Figures~\ref{fig:refinement-pipeline} and~\ref{fig:refinement-pipeline-perfect}.


\subsection{Comparison with \cite{DBLP:conf/aicol/DragoniVRG17}}
\label{comparison-with-the-literature}

In this section, we compare our experimental findings with the results presented by~\cite{DBLP:conf/aicol/DragoniVRG17}. 
To ensure methodological consistency, we restrict the comparison to the \textit{lower branch evaluation} reported in their study.
For this purpose, we employ the standard metrics of {\em Precision}, {\em Recall}, and their harmonic mean, the {F1-score}, defined as follows:
\[
\renewcommand{\arraystretch}{2.25}
\begin{array}{lcl}
\text{Precision} &=& \displaystyle\frac{\text{TP}}{\text{TP} + \text{FP}} = \displaystyle\frac{\text{Matched Items}}{\text{Generated Items}}\\
\text{Recall} &=&  \displaystyle\frac{\text{TP}}{\text{TP} + \text{FN}} = \displaystyle\frac{\text{Matched Items}}{\text{Gold Standard Items}}\\
\text{F1} & = & 2 \cdot \displaystyle\frac{\text{Precision} \cdot \text{Recall}}{\text{Precision} + \text{Recall}}
\end{array}
\]
In this context, {\em Items} refers to either atoms or rules, depending on the evaluation. True positives (TP) are items generated by the LLM that are also found in the gold standard. False positives (FP) are generated items not present in the gold standard. False negatives (FN) are not explicitly observed. The denominator in the classical recall formula, $\text{TP} + \text{FN}$, serves as a proxy for the total number of positives in the labeled set. However, in our case, the gold standard itself is the labeled set and contains only positives. Therefore, the total number of positives is directly known and equals the number of items in the gold standard.

Our analysis reveals an immediate discrepancy: \cite{DBLP:conf/aicol/DragoniVRG17} report 65 terms and 36 rules in the gold standard, whereas we identified 69 terms and 52 rules across Sections 8.2.1(a)–-8.2.1(c) in the gold standard.

Importantly, in the following analysis, we evaluate the precision and recall of the formalizations produced by the LLMs based on our counts.

\subsubsection{Evaluation of Term Identification}

The first level of comparison involves the correct identification of legal terms (referred to as {\em atoms}).
\cite{DBLP:conf/aicol/DragoniVRG17} reported the following: \enquote{\em The gold standard contains 65 terms extracted by the analysts; our system is able to extract 59 terms whose 49 are correct. Therefore, the obtained recall is $90.78\%$ and the precision is $83.05\%$, with a F-Measure of $86.74\%$.} However, it should be noted that these figures are inconsistent. While the precision has been correctly calculated, the recall based on the reported numbers actually amounts to $49/65 = 75.38\%$ (assuming the stated 65 atoms) rather than the claimed $90.78\%$. Moreover, this discrepancy also leads to a recalculated F1-Measure of $79.03\%$ instead of the reported $86.74\%$.
Table~\ref{tab:term-identification} summarizes the precision, recall and F1-score values achieved across various configurations and models in our study and compares them with the corrected values reported by \cite{DBLP:conf/aicol/DragoniVRG17}.
The highest precision was still achieved by \cite{DBLP:conf/aicol/DragoniVRG17}, while the second-best result was obtained by GPT-4o when formalizing all law snippets in a single step. 
The highest recall was achieved with the additional refinement step described in Section~\ref{refinement-step} when GPT-4.1 was employed as the initial model for baseline formalization, whereas the highest F1-score was obtained under the same refinement approach when DeepSeek-R1 (0528) was used in the first stage.

\newcommand{\resultsfromdragonietal}{\multirow{2}{*}{\shortstack[c]{Results from \\ \cite{DBLP:conf/aicol/DragoniVRG17}}} }
\newcommand{\comparisonexperimentcoi}{\multirow{5}{*}{\shortstack[c]{Simple CoI \\ (Section \ref{determining-the-best-prompt})}} }
\newcommand{\comparisonexperimentsingleinteraction}{\multirow{4}{*}{\shortstack[c]{Single \\ Interaction \\ (Section \ref{single-interaction})}}}
\newcommand{\comparisonexperimentft}{\multirow{4}{*}{\shortstack[c]{Fine-tuned \\ after Epoch 1 \\ (Section \ref{finetuning-the-models})}}}
\newcommand{\comparisonexperimenttwostage}{\multirow{4}{*}{\shortstack[c]{Refinement Step \\ w. Claude Sonnet 4 \\ (Section \ref{refinement-step})}}}
\newcommand{\highlightcell}[1]{$\mathbf{#1}$}

\begin{table*}[h]
    \centering
    \small
    \caption{Precision, Recall and F1-score for Term Identification}
    \label{tab:term-identification}

    \begin{threeparttable}
    \begin{tabular}{cccccc}
        \toprule
        Experiment & Model & \makecell{Matched/ \\ Generated \\ Atoms} & Precision & Recall & F1 \\\midrule

        \resultsfromdragonietal
        & reported 65 atoms & \multirow{2}{*}{$49/59$} & $83.05\%$ & $75.38\%$\tnote{1} & $0.79$\tnote{1} \\
        & counted 69 atoms\tnote{2} & & \highlightcell{83.05\%} & $71.01\%$\tnote{2} & $0.77$\tnote{2} \\\midrule
        
        \comparisonexperimentcoi
        & GPT-4.1 & $59/91$ & $64.84\%$ & $85.51\%$ & $0.74$ \\
        & Cl. Sonnet 4 (ET) & $58/88$ & $65.91\%$ & $84.06\%$ & $0.74$ \\
        & DeepSeek-R1 (0528) & $59/80$ & $73.75\%$ & $85.51\%$ & $0.79$ \\
        & o3 & $58/93$ & $62.37\%$ & $84.06\%$ & $0.72$ \\
        & o4-mini & $58/96$ & $60.42\%$ & $84.06\%$ & $0.70$ \\
        & GPT-4o & $55/77$ & $71.43\%$ & $79.71\%$ & $0.75$ \\\midrule

        \comparisonexperimentsingleinteraction
        & GPT-4.1 & $53/68$ & $77.94\%$ & $76.81\%$ & $0.77$ \\
        & DeepSeek-R1 (0528) & $55/68$ & $80.88\%$ & $79.71\%$ & $0.80$\\
        & o4-mini & $51/64$ & $79.69\%$ & $73.91\%$ & $0.77$ \\
        & GPT-4o & $53/64$ & $82.81\%$ & $76.81\%$ & $0.80$ \\\midrule

        \comparisonexperimentft
        & GPT-4.1 (Config 2) & $55/77$ & $71.43\%$ & $79.71\%$ & $0.75$ \\
        & GPT-4.1 (Config 3) & $53/83$ & $63.86\%$ & $76.81\%$ & $0.70$ \\
        & GPT-4o (Config 2) & $56/71$ & $78.87\%$ & $81.16\%$ & $0.80$ \\
        & GPT-4o (Config 3) & $57/72$ & $79.17\%$ & $82.61\%$ & $0.81$ \\\midrule

        \comparisonexperimenttwostage
        & GPT-4.1 & $60/85$ & $70.59\%$ & \highlightcell{86.96\%} & $0.78$ \\
        & Cl. Sonnet 4 (ET) & $58/84$ & $69.05\%$ & $84.06\%$ & $0.76$ \\
        & DeepSeek-R1 (0528) & $58/73$ & $79.45\%$ & $84.06\%$ & \highlightcell{0.82} \\
        & o3 & $58/74$ & $78.38\%$ & $84.06\%$ & $0.81$ \\
        \bottomrule
    \end{tabular}
    \begin{tablenotes}
        \item[1] We recalculated the Recall and F1-score, as the values reported in \cite{DBLP:conf/aicol/DragoniVRG17} are inconsistent. Based on 65 atoms in the gold standard and 49 correct out of 59 generated atoms, the Precision is $75.38\%$ instead of $90.78\%$, resulting in an F1-score of $0.79$ rather than $0.86$.
        \item[2] We counted 69 atoms in the gold standard as opposed to the 65 reported in \cite{DBLP:conf/aicol/DragoniVRG17}. Therefore, we provide calculations based on both figures.
    \end{tablenotes}
    \end{threeparttable}
\end{table*}

\subsubsection{Evaluation of Deontic Annotation Accuracy}
The second dimension of analysis concerns the accurate assignment of deontic modality (i.e., obligation, permission, or prohibition).
In the benchmark study, $47$ of $49$ correctly identified atoms were accurately annotated, yielding a deontic annotation precision of $95.92\%$.
In contrast, across all experiments conducted in this work, $100\%$ of atoms -- both correctly identified atoms and such without a counterpart in the gold standard -- were annotated with the correct deontic label.
Thus, the deontic annotation precision in our experiments is consistently $100\%$.

\subsubsection{Identification of Rule Counterparts}
The third criterion evaluates the number of generated rules that have a semantically corresponding rule in the gold standard.
Following the method defined in~\cite{DBLP:conf/aicol/DragoniVRG17}, a rule is considered a {\em counterpart} if there is a semantic match in its consequent with a rule in the manually curated set of the gold standard.
In their evaluation, $33$ out of $41$ generated rules had counterparts among the reported $36$ gold standard rules, resulting in a precision of $80.49\%$ and a recall of $91.67\%$.
The metrics evaluation computed with the reported $36$ gold standard rules, the evaluation corrected with the $52$ rules we counted, and the results from our models are all presented in Table~\ref{tab:counterpart-identification}.

The highest precision (87.18\%) was achieved by DeepSeek-R1 (0528) when formalizing all law snippets in a single interaction. The highest recall (82.69\%) was attained by GPT-4.1 and o3 in Section~\ref{determining-the-best-prompt}, as well as by using GPT-4.1 as first model in Section~\ref{refinement-step}. The highest F1-score was obtained with a fine-tuned version of GPT-4.1, as well as by using o3 as model for the initial formalization in Section~\ref{refinement-step}.

\begin{table*}[h]
    \centering
    \small
    \caption{Precision, Recall and F1-score for Counterpart Identification}
    \label{tab:counterpart-identification}
    \begin{threeparttable}
    \begin{tabular}{cccccc}
        \toprule
        Experiment & Model & \makecell{Matched/ \\ Generated \\ Rules} & Precision & Recall & F1 \\\midrule

        \resultsfromdragonietal
        & reported 36 rules & \multirow{2}{*}{33/41} & $80.49\%$ & $91.67\%$ & $0.86$ \\
        &  counted 52 rules\tnote{1} & & $80.49\%$ & $63.46\%$\tnote{1} & $0.71$\tnote{1}  \\\midrule
        
        \comparisonexperimentcoi
        & GPT-4.1 & $43/56$ & $76.79\%$ & \highlightcell{82.69\%} & $0.80$ \\
        & Cl. Sonnet 4 (ET) & $41/50$ & $82.00\%$ & $78.85\%$ & $0.80$ \\
        & DeepSeek-R1 (0528) & $36/43$ & $83.72\%$ & $69.23\%$ & $0.76$ \\
        & o3 & $43/57$ & $75.44\%$ & \highlightcell{82.69\%} & $0.79$ \\
        & o4-mini & $37/51$ & $72.55\%$ & $71.15\%$ & $0.72$ \\
        & GPT-4o & $39/47$ & $82.98\%$ & $75.00\%$ & $0.79$ \\\midrule

        \comparisonexperimentsingleinteraction
        & GPT-4.1 & $30/37$ & $81.08\%$ & $57.69\%$ & $0.67$ \\
        & DeepSeek-R1 (0528) & $34/39$ & \highlightcell{87.18\%} & $65.38\%$ & $0.75$ \\
        & o4-mini & $40/47$ & $85.11\%$ & $76.92\%$ & $0.81$ \\
        & GPT-4o & $29/36$ & $80.56\%$ & $55.77\%$ & $0.66$ \\\midrule

        \comparisonexperimentft
        & GPT-4.1 (Config 2) & $38/45$ & $84.44\%$ & $73.08\%$ & $0.78$ \\
        & GPT-4.1 (Config 3) & $40/46$ & $86.96\%$ & $76.92\%$ & \highlightcell{0.82} \\
        & GPT-4o (Config 2) & $38/45$ & $84.44\%$ & $73.08\%$ & $0.78$ \\
        & GPT-4o (Config 3) & $38/45$ & $84.44\%$ & $73.08\%$ & $0.78$ \\\midrule

        \comparisonexperimenttwostage
        & GPT-4.1 & $43/56$ & $76.79\%$ & \highlightcell{82.69\%} & $0.80$ \\
        & Cl. Sonnet 4 (ET) & $40/50$ & $80.00\%$ & $76.92\%$ & $0.78$ \\
        & DeepSeek-R1 (0528) & $36/43$ & $83.72\%$ & $69.23\%$ & $0.76$ \\
        & o3 & $42/51$ & $82.35\%$ & $80.77\%$ & \highlightcell{0.82} \\
        \bottomrule
    \end{tabular}
    \begin{tablenotes}
        \item[1] {\bf Note.} We counted 52 rules in the gold standard as opposed to the 36 reported in \cite{DBLP:conf/aicol/DragoniVRG17}. Therefore, we present both sets of metric values: those originally reported and those recalculated based on our count. For the purpose of comparing with our LLM-based approach, we use the recalculated values as reference. 
    \end{tablenotes}
    \end{threeparttable}
\end{table*}

\subsubsection{Evaluation of Full Rule Correspondence}
Finally, we assess the degree of {\em full correspondence}, where a rule in the generated set semantically matches the gold standard in both antecedents and consequent.
Dragoni et al.\ report that $24$ rules in their generated set fully matched semantically their counterparts in the gold standard, with a resulting precision of $58.54\%$ and recall of $66.67\%$.
The corresponding results from our evaluation are displayed in Table~\ref{tab:full-correspondence-identification}.

The highest precision ($76.92\%$) was achieved by DeepSeek-R1 (0528) in Section \ref{single-interaction}, while the highest recall ($75\%$) was achieved by using GPT-4.1 as first model in Section \ref{refinement-step}. Moreover, the highest F1-score ($0.74$) was achieved by refining the output of o3 in a two-stage pipeline.

\begin{table*}[h]
    \centering
    \small
    \caption{Precision, Recall and F1-score for Full Rule Correspondence}
    \label{tab:full-correspondence-identification}
    \begin{threeparttable}
    \begin{tabular}{cccccc}
        \toprule
        Experiment & Model & \makecell{Matched/ \\ Generated \\ Rules} & Precision & Recall & F1 \\\midrule

        \resultsfromdragonietal
        & reported 36 rules & \multirow{2}{*}{$24/41$} & $58.54\%$ & $66.67\%$ & $0.62$ \\
        & counted 52 rules\tnote{1} & & $58.54\%$ & $46.15\%$\tnote{1} & $0.52$\tnote{1} \\\midrule
        
        \comparisonexperimentcoi
        & GPT-4.1 & $37/56$ & $66.07\%$ & $71.15\%$ & $0.69$ \\
        & Cl. Sonnet 4 (ET) & $36/50$ & $72.00\%$ & $69.23\%$ & $0.71$ \\
        & DeepSeek-R1 (0528) & $31/43$ & $72.09\%$ & $59.62\%$ & $0.65$ \\
        & o3 & $37/57$ & $64.91\%$ & $71.15\%$ & $0.68$ \\
        & o4-mini & $30/51$ & $58.82\%$ & $57.69\%$ & $0.58$ \\
        & GPT-4o & $30/47$ & $63.83\%$ & $57.69\%$ & $0.61$ \\\midrule

        \comparisonexperimentsingleinteraction
        & GPT-4.1 & $28/37$ & $75.68\%$ & $53.85\%$ & $0.63$ \\
    & DeepSeek-R1 (0528) & $30/39$ & \highlightcell{76.92\%} & $57.69\%$ & $0.66$ \\
        & o4-mini & $36/47$ & $76.60\%$ & $69.23\%$ & $0.73$ \\
        & GPT-4o & $27/36$ & $75.00\%$ & $51.92\%$ & $0.61$ \\\midrule

        \comparisonexperimentft
        & GPT-4.1 (Config 2) & $33/45$ & $73.33\%$ & $63.46\%$ & $0.68$ \\
        & GPT-4.1 (Config 3) & $33/46$ & $71.74\%$ & $63.46\%$ & $0.67$ \\
        & GPT-4o (Config 2) & $34/45$ & $75.56\%$ & $65.38\%$ & $0.70$ \\
        & GPT-4o (Config 3) & $33/45$ & $73.33\%$ & $63.46\%$ & $0.68$ \\\midrule

        \comparisonexperimenttwostage
        & GPT-4.1 & $39/56$ & $69.64\%$ & \highlightcell{75.00\%} & $0.72$ \\
        & Cl. Sonnet 4 (ET) & $36/50$ & $72.00\%$ & $69.23\%$ & $0.71$ \\
        & DeepSeek-R1 (0528) & $32/43$ & $74.42\%$ & $61.54\%$ & $0.67$ \\
        & o3 & $38/51$ & $74.51\%$ & $73.08\%$ & \highlightcell{0.74}\\
        \bottomrule
    \end{tabular}
    \begin{tablenotes}
        \item[1] Note that we counted 52 rules in the gold standard as opposed to the 36 reported in \cite{DBLP:conf/aicol/DragoniVRG17}. Therefore, we provide calculations based on both figures. 
    \end{tablenotes}
    \end{threeparttable}
\end{table*}

\subsubsection{Limitations of the Compared Evaluation Metrics}
\label{limitations-compared-evaluation-metrics}
While the evaluation framework used by Dragoni et al.\ has certain benefits -- such as penalizing insufficient reuse of atoms across legal clauses -- it also presents some limitations.

A key issue lies in the one-to-one rule mapping constraint: each rule in the gold standard can be matched to at most one rule in the generated set and vice versa.
This restriction becomes problematic in cases where an LLM produces multiple valid rules which all together are equivalent to a single rule in the gold standard.
In such scenarios, semantically correct rules are penalized due to a lack of corresponding entries in the gold standard.

For instance, in the formalization of law snippet 8.2.1.a.i, the gold standard specifies only three rules, while the LLM-generated formalizations typically consist of six rules, offering a more detailed representation.
Nonetheless, these additional rules are considered false positives in the evaluation, thus lowering precision.

The same limitation applies to term identification: valid atoms that are not present in the gold standard reduce the measured precision, even if their extraction is semantically justified.

An additional shortcoming is that LLMs are penalized for formalizing additional information from the law text that is not represented in the gold standard. Consider, for example, law snippet 8.2.1.b, which contains the following sentence: 

\begin{quote}
    ''If a Consumer tells the Supplier that they are dissatisfied with the timeframes that apply to the management of a Complaint or seek to have a Complaint treated as an Urgent Complaint, the Supplier must tell the Consumer about the Supplier's internal prioritisation and internal escalation processes.``
\end{quote}

In the gold standard, this clause is formalized as:

\begin{ruletable}
    \tablecenter{customerDissatisfiedTimeframe(X) $\Rightarrow$ [O] informInternalPrioritisation(X)} \\\addlinespace
    \tablecenter{customerDissatisfiedTimeframe(X) $\Rightarrow$ [O] informInternalEscalationProcess(X)} \\\addlinespace
    \tablecenter{escalation(X), internalPrioritisation(X) $\Rightarrow$ [O] informExternalDisputeResolution(X)}
\end{ruletable}

However, some LLMs produced the following additional formalizations:

\begin{ruletable}
    \tablecenter{complaint(X), consumerRequestsUrgent(X) $\Rightarrow$ [O] informInternalPrioritisation(X)}\\
    \addlinespace
    \tablecenter{complaint(X), consumerRequestsUrgent(X) $\Rightarrow$ [O] informInternalEscalation(X)}
\end{ruletable}

Although these rules are arguably justified by the legal text, they are penalized under the evaluation framework due to their absence from the gold standard.
Thus, the metric fails to distinguish between semantically valid additions and hallucinated additions, undermining its reliability in assessing true model performance.

\section{Limitations}
\label{sec:limitations}

We describe both the intrinsic limitations and the technical challenges we encountered.
\subsection{Legal Interpretation}
A major challenge for the formal encoding of legal documents is that each encoding is an interpretation, and the gold standard should correspond to the authentic interpretation. However, in some jurisdictions, it will not be possible to have a true gold standard. The gold standard would correspond to the authentic interpretation of the legal provision, and the only authority able to provide an authentic interpretation is the judiciary. Moreover, this is possible only for cases disputed in court, and it would be limited to the provisions effectively used in the legal proceeding. 
The second issue is that a legal interpretation depends on the understanding of the legal intent, legal context and the encoding style of the coders.    \cite{ail2022:Alice} reports on an empirical experiment where three (experienced) coders were asked to model in DDL the same set of legal provisions (from the Australian Copyright Act). The experiment had two phases; in the first phase, the coders did the encoding fully independently. In the second phase, the coders agreed on a common set of terms and then encoded independently the provisions as rules. In the first phase, the degree of agreement varied from 0\% to 10\% for terms, and 0\% of rules using a perfect match, and around 50\% for terms and 3\% on rules with a semantics correspondence. In phase two, the term agreement was between 30\% and 56\% for the full correspondence and 85\% for semantic correspondence; the rule similarity ranged from 10\% to 30\% for full correspondence and 26\% to 53\% with semantic correspondence. 

\subsection{Inter-Paragraph References} 
A common strategy for managing the complexity of legal documents is the use of references, which may be  internal (linking to other sections within the same document) or external (referring to other documents) \citep{DBLP:conf/icail/GovernatoriO21}. 
Accordingly, the dataset used in this study included occasional references between different legal paragraphs.
For example, law snippet 8.2.1.a.xiv mandates that a complaint must only be closed \enquote{\textit{with the consent of the Consumer or former Customer or if clauses 8.2.1(c),(d) or (e) below have been complied with}}.

The LLMs generated suitable atoms such as \texttt{clause8.2.1.c.complied(X)}, but these were not reused in the formalizations of the referenced paragraphs (i.e., 8.2.1(c), (d), or (e)).
As a consequence, the preconditions set in the formalization of law snippet 8.2.1(a)(xiv) could not be met, rendering this rule ineffective within the formal system. 

This problem persisted even when formalizing all law snippets in a single interaction or when applying a refinement step. 
This limitation suggests that prompt engineering alone is insufficient to fully address the challenge of reference resolution.
Instead, it requires the incorporation of additional procedural components into the methodology.
One possible solution is a refinement phase following the initial generation process, designed to ensure semantic coherence across references.

\section{Conclusion and Future Work}
\label{sec:conclusion-and-future-work}

This paper demonstrated that LLMs can be effectively leveraged to transform legal norms into formal DDL rules, achieving performance that is in most respects superior to the approach proposed in~\cite{DBLP:conf/aicol/DragoniVRG17}. 
Based on recalculated metrics that reflect the actual number of rules and atoms in the gold standard, our method outperforms Dragoni et al. across nearly all evaluation settings. The only exception concerns precision in the term identification task, where Dragoni et al. attains a higher score.

Our evaluation indicates that prompt engineering -- particularly the use of few-shot learning combined with Chain-of-Instructions -- yields promising results. 
By contrast, neither the direct formalization of all law snippets in a single step nor fine-tuning led to measurable performance gains.
Moreover, decomposing the task into a two-step pipeline consisting of atom extraction followed by rule generation resulted in a decline in output quality.
A modest improvement was, however, observed when employing a two-step pipeline with an additional refinement stage after rule generation.
This refinement step contributed, among other aspects, to greater consistency in the use of atom names across law snippets.

Future work includes integrating active learning and expert-in-the-loop feedback to continuously refine LLM outputs.
Expanding the domain beyond TCP Code and adapting the pipeline to multilingual legal corpora could further validate the generalizability of our approach.
Moreover, the formalization of the superiority relationship -- currently omitted due to limited occurrences in the dataset -- deserves further investigation, potentially via prompt engineering or a dedicated pipeline stage.
Finally, embedding these methods in end-user tools for compliance and regulatory auditing represents a practical next step.

\bmhead{Acknowledgments}
    This work has been partially funded by the Vienna Science and Technology Fund (WWTF) [Grant ID: 10.47379/ICT23030].

\bibliography{bibliography}

\end{document}